\documentclass[a4paper,twoside]{article}

\usepackage{epsfig}
\usepackage{subcaption}
\usepackage{calc}
\usepackage{amssymb}
\usepackage{amstext}
\usepackage{amsmath}
\usepackage{amsthm}
\usepackage{multicol}
\usepackage{pslatex}
\usepackage{apalike}
\usepackage{hyperref}

\usepackage{xspace}
\newcommand{\di}{A2D2\xspace}

\newcommand{\adas}{ADAS\xspace}
\newcommand{\lidar}{LiDAR\xspace}
\newcommand{\kitti}{KITTI\xspace}

\usepackage[inline]{enumitem}
\usepackage{xcolor}
\newcommand{\RM}[1]{}
\newcommand{\MS}[1]{}
\newcommand{\BE}[1]{}

\usepackage{tikz}
\usepackage{pgfplots}
\usetikzlibrary{spy, calc}
\newif\ifblackandwhitecycle
\gdef\patternnumber{0}

\pgfkeys{/tikz/.cd,
	zoombox paths/.style={
		draw=orange,
		very thick
	},
	black and white/.is choice,
	black and white/.default=static,
	black and white/static/.style={ 
		draw=white,   
		zoombox paths/.append style={
			draw=white,
			postaction={
				draw=black,
				loosely dashed
			}
		}
	},
	black and white/static/.code={
		\gdef\patternnumber{1}
	},
	black and white/cycle/.code={
		\blackandwhitecycletrue
		\gdef\patternnumber{1}
	},
	black and white pattern/.is choice,
	black and white pattern/0/.style={},
	black and white pattern/1/.style={    
		draw=white,
		postaction={
			draw=black,
			dash pattern=on 2pt off 2pt
		}
	},
	black and white pattern/2/.style={    
		draw=white,
		postaction={
			draw=black,
			dash pattern=on 4pt off 4pt
		}
	},
	black and white pattern/3/.style={    
		draw=white,
		postaction={
			draw=black,
			dash pattern=on 4pt off 4pt on 1pt off 4pt
		}
	},
	black and white pattern/4/.style={    
		draw=white,
		postaction={
			draw=black,
			dash pattern=on 4pt off 2pt on 2 pt off 2pt on 2 pt off 2pt
		}
	},
	zoomboxarray inner gap/.initial=5pt,
	zoomboxarray columns/.initial=2,
	zoomboxarray rows/.initial=2,
	subfigurename/.initial={},
	figurename/.initial={zoombox},
	zoomboxarray/.style={
		execute at begin picture={
			\begin{scope}[
				spy using outlines={%
					zoombox paths,
					width=\imagewidth / \pgfkeysvalueof{/tikz/zoomboxarray columns} - (\pgfkeysvalueof{/tikz/zoomboxarray columns} - 1) / \pgfkeysvalueof{/tikz/zoomboxarray columns} * \pgfkeysvalueof{/tikz/zoomboxarray inner gap} -\pgflinewidth,
					height=\imageheight / \pgfkeysvalueof{/tikz/zoomboxarray rows} - (\pgfkeysvalueof{/tikz/zoomboxarray rows} - 1) / \pgfkeysvalueof{/tikz/zoomboxarray rows} * \pgfkeysvalueof{/tikz/zoomboxarray inner gap}-\pgflinewidth,
					magnification=3,
					every spy on node/.style={
						zoombox paths
					},
					every spy in node/.style={
						zoombox paths
					}
				}
				]
			},
			execute at end picture={
			\end{scope}
			\gdef\patternnumber{0}
		},
		spymargin/.initial=0.5em,
		zoomboxes xshift/.initial=1,
		zoomboxes right/.code=\pgfkeys{/tikz/zoomboxes xshift=1},
		zoomboxes left/.code=\pgfkeys{/tikz/zoomboxes xshift=-1},
		zoomboxes yshift/.initial=0,
		zoomboxes above/.code={
			\pgfkeys{/tikz/zoomboxes yshift=1},
			\pgfkeys{/tikz/zoomboxes xshift=0}
		},
		zoomboxes below/.code={
			\pgfkeys{/tikz/zoomboxes yshift=-1},
			\pgfkeys{/tikz/zoomboxes xshift=0}
		},
		caption margin/.initial=4ex,
	},
	adjust caption spacing/.code={},
	image container/.style={
		inner sep=0pt,
		at=(image.north),
		anchor=north,
		adjust caption spacing
	},
	zoomboxes container/.style={
		inner sep=0pt,
		at=(image.north),
		anchor=north,
		name=zoomboxes container,
		xshift=\pgfkeysvalueof{/tikz/zoomboxes xshift}*(\imagewidth+\pgfkeysvalueof{/tikz/spymargin}),
		yshift=\pgfkeysvalueof{/tikz/zoomboxes yshift}*(\imageheight+\pgfkeysvalueof{/tikz/spymargin}+\pgfkeysvalueof{/tikz/caption margin}),
		adjust caption spacing
	},
	calculate dimensions/.code={
		\pgfpointdiff{\pgfpointanchor{image}{south west} }{\pgfpointanchor{image}{north east} }
		\pgfgetlastxy{\imagewidth}{\imageheight}
		\global\let\imagewidth=\imagewidth
		\global\let\imageheight=\imageheight
		\gdef\columncount{1}
		\gdef\rowcount{1}
		
	},
	image node/.style={
		inner sep=0pt,
		name=image,
		anchor=south west,
		append after command={
			[calculate dimensions]
			node [image container,subfigurename=\pgfkeysvalueof{/tikz/figurename}-image] {\phantomimage}
			node [zoomboxes container,subfigurename=\pgfkeysvalueof{/tikz/figurename}-zoom] {\phantomimage}
		}
	},
	color code/.style={
		zoombox paths/.append style={draw=#1}
	},
	connect zoomboxes/.style={
		spy connection path={\draw[draw=none,zoombox paths] (tikzspyonnode) -- (tikzspyinnode);}
	},
	help grid code/.code={
		\begin{scope}[
			x={(image.south east)},
			y={(image.north west)},
			font=\footnotesize,
			help lines,
			overlay
			]
			\foreach \x in {0,1,...,9} { 
				\draw(\x/10,0) -- (\x/10,1);
				\node [anchor=north] at (\x/10,0) {0.\x};
			}
			\foreach \y in {0,1,...,9} {
				\draw(0,\y/10) -- (1,\y/10);                        \node [anchor=east] at (0,\y/10) {0.\y};
			}
		\end{scope}    
	},
	help grid/.style={
		append after command={
			[help grid code]
		}
	},
}

\newcommand\phantomimage{%
	\phantom{%
		\rule{\imagewidth}{\imageheight}%
	}%
}
\newcommand\zoombox[2][]{
	\begin{scope}[zoombox paths]
		\pgfmathsetmacro\xpos{
			(\columncount-1)*(\imagewidth / \pgfkeysvalueof{/tikz/zoomboxarray columns} + \pgfkeysvalueof{/tikz/zoomboxarray inner gap} / \pgfkeysvalueof{/tikz/zoomboxarray columns} ) + \pgflinewidth
		}
		\pgfmathsetmacro\ypos{
			(\rowcount-1)*( \imageheight / \pgfkeysvalueof{/tikz/zoomboxarray rows} + \pgfkeysvalueof{/tikz/zoomboxarray inner gap} / \pgfkeysvalueof{/tikz/zoomboxarray rows} ) + 0.5*\pgflinewidth
		}
		\edef\dospy{\noexpand\spy [
			#1,
			zoombox paths/.append style={
				black and white pattern=\patternnumber
			},
			every spy on node/.append style={#1},
			x=\imagewidth,
			y=\imageheight
			] on (#2) in node [anchor=north west] at ($(zoomboxes container.north west)+(\xpos pt,-\ypos pt)$);}
		\dospy
		\pgfmathtruncatemacro\pgfmathresult{ifthenelse(\columncount==\pgfkeysvalueof{/tikz/zoomboxarray columns},\rowcount+1,\rowcount)}
		\global\let\rowcount=\pgfmathresult
		\pgfmathtruncatemacro\pgfmathresult{ifthenelse(\columncount==\pgfkeysvalueof{/tikz/zoomboxarray columns},1,\columncount+1)}
		\global\let\columncount=\pgfmathresult
		\ifblackandwhitecycle
		\pgfmathtruncatemacro{\newpatternnumber}{\patternnumber+1}
		\global\edef\patternnumber{\newpatternnumber}
		\fi
	\end{scope}
}

\usepackage{SCITEPRESS}     % Please add other packages that you may need BEFORE the SCITEPRESS.sty package.

\begin{document}

\title{A Lightweight Machine Learning Pipeline for \lidar-simulation}

\author{
\authorname{Richard Marcus \sup{1}\orcidAuthor{0000-0002-6601-6457},
Niklas Knoop \sup{2},
Bernhard Egger \sup{1}\orcidAuthor{0000-0002-4736-2397}
and Marc Stamminger \sup{1}\orcidAuthor{0000-0001-8699-3442}
}
\affiliation{\sup{1} Chair of Visual Computing, Friedrich-Alexander-Universität Erlangen-Nürnberg, Germany}
\affiliation{\sup{2} Elektronische Fahrwerksysteme GmbH, Germany}
\email{\{richard.marcus, bernhard.egger, marc.stamminger\}@fau.de, niklas.knoop@efs-auto.de}
}


\abstract{
Virtual testing is a crucial task to ensure safety in autonomous driving, and sensor simulation is an important task in this domain.
Most current \lidar simulations are very simplistic and are mainly used to perform initial tests, while the majority of insights are gathered on the road.
In this paper, we propose a lightweight approach for more realistic \lidar simulation that learns a real sensor's behavior from test drive data and transforms this to the virtual domain.
The central idea is to cast the simulation into an image-to-image translation problem.
We train our pix2pix based architecture on two real world data sets, namely the popular \kitti data set and the Audi Autonomous Driving Dataset which provide both, RGB and \lidar images.
We apply this network on synthetic renderings and show that it generalizes sufficiently from real images to simulated images.
This strategy enables to skip the sensor-specific, expensive and complex \lidar physics simulation in our synthetic world and avoids oversimplification and a large domain-gap through the clean synthetic environment.
%This enables generic import of the pipeline into simulation environments, which can use the simulated \lidar data as input for Advanced Driver Assistance Systems (\adas) and AD.
}

\onecolumn \maketitle \normalsize \setcounter{footnote}{0} \vfill

\section{\uppercase{Introduction}}
\label{sec:introduction}
%\RM{wenn noch Zeit ist, mehr auf Bilder oder sections referenzieren?}
Even though Autonomous Driving (AD) and Advanced Driver Assistance Systems (\adas) have been a major research areas for more than a decade, the adoption into practical driving systems is dragging on.
With the lacking capabilities of current algorithms to adapt to unforeseen situations, a big challenge is testing the performance of such systems safely. 
%Even with test drivers, system failures have often lead to real world accidents in routine scenarios.
%Beyond that, it is almost impossible to test dangerous manoeuvres such as evading obstacles or driving under harsh weather conditions. 
%Therefore, it is of utmost importance to obtain virtual testing methods that accurately reflect the real world and consequently can be used to transfer insights of virtual \adas to the real world.
A key part to achieve this vision is the realistic simulation of the car sensors such as cameras, \lidar or RADAR.
Simulating such sensors in a virtual environment is costly, in particular if the virtual sensors should suffer from the same physical limitations and imperfections as their real counterpart.

%A typical approach is to {\em simulate} the sensor as precisely as possible, which requires complex and costly physical simulation as well as extensive data about the sensor and the environment.
A typical approach are physical based simulations, which require extensive data about the sensor specifics and a careful implementation.
A more desirable option is to {\em learn} the behavior of a particular sensor from recorded real test drive data and to transfer this to a virtual sensor.
The process should be mostly automatic, in order to easily adapt the simulation to new models and types of sensors.

The core idea of this paper is to tackle the problem as an image-to-image translation task.
In particular, we consider the simulation of \lidar sensors, which send out light beams in a uniform cylindrical grid and measure the time of flight and thus the distance of the scene point visible in this particular direction.
If there is no hit point (sky), the hit point absorbs the light (dark surfaces), or does not reflect back (mirroring surface) or multiple reflections interfere, no measurement is returned.
Of course, the possibilities to learn sensor behavior strongly depend on the available test drive (and thus ground truth data).

\begin{figure}[ht]
	\centering
	\includegraphics[width=.95\linewidth]{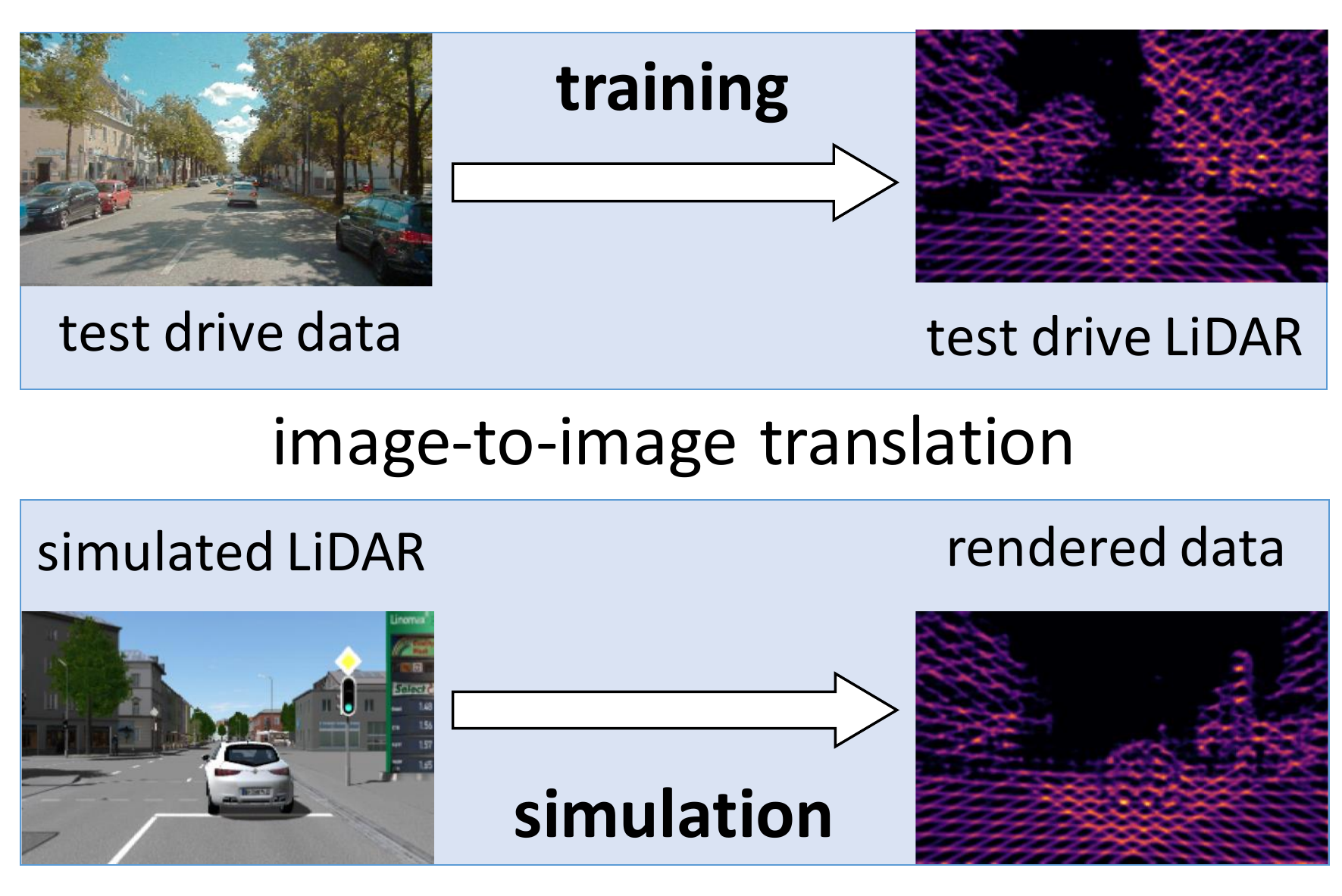}
	\caption{We learn the translation from real world RGB to \lidar data from test drives. We use this learned mapping to derive \lidar output from rendered RGB images.}
	\label{fig:idea}
\end{figure}

Relevant for our purpose are a RGB camera stream and a synchronized stream of LiDAR points. %, some data sets also provide hand-annotated segmentation masks. 
Our concept is shown in Fig.~\ref{fig:idea}:
from recorded test drive data, we learn the transfer from RGB camera images to \lidar images using the Pix2Pix architecture \cite{isola_image--image_2018}.
%Later, during virtual testing, we use the thus trained network to transform a rendered RGB image to the corresponding \lidar response (right).
Fig.~\ref{fig:idea} already shows a potential challenge of the approach: real and simulated RGB images still have significantly different characteristics.
It is unclear, how much this gap between real and simulated imagery influences the outcome in the simulation.
We thus examine other input image types, e.g.~depth images, or segmentation images, where
%For such simpler images
the real-2-sim gap might be smaller.%, yet test drive data does not provide these images, so we derive them during training.

In this study, we explore variants of this idea and show that it is possible to learn a plausible LiDAR simulation solely from test drive data, in our case KITTI and A2D2.
Our simulation runs in real-time and can thus be integrated in a virtual test drive system.
We compare the different modalities of input images and ablate which are most suitable for our learned \lidar simulation strategy.
We evaluate our idea by examining how well ground truth \lidar data can be reproduced with our approach, and how well we can predict real \lidar measurements from synthetic rendered data using the second revision~\cite{cabon2020vkitti2} of the Virtual KITTI data set (VKITTI).

\section{\uppercase{Related Work}}

\paragraph{Simulation Environments}
Many modern simulation environments like LGSVL~\cite{simulator_lgsvl_nodate}, CARLA~\cite{dosovitskiy_carla_2017} or Intel Airsim~\cite{shah_airsim_2017} build on game engines to offer high quality graphics and profit from their development and communities.
Additionally, there are different \adas integrations, sensor configurations and simulation settings, e.g.~weather or traffic, to support many use cases of automotive simulation.
Other tools like VTD~\cite{VTD} or IPG Carmaker~\cite{ipg_carmaker_2020} focus on supporting their own software stack and promise a more complete solution.

%While there are still interfaces to external tools, other solutions generally offer more visualization settings.
These environments also offer sensor simulations.
Yet, existing \lidar simulations are usually simplistic and not sufficient to test \adas or autonomous vehicles, so we see potential for our approach here.
To integrate our approach into these systems, we only need to get access to rendered RGB images, their depth buffer, or segmentation masks, so the integration of a new module that transfers these images to \lidar sensor output is very simple.

%All of these allow users to integrate own machine learning based sensor simulation that generates images during runtime for further processing.
%While the quality of the renderings can affect the results drastically, simulation tools can always fall back on visualizing the depth or semantic segmentations when the photorealism proves to be lacking.

Nvidia DRIVE Sim~\cite{nvidia_drivesim} is different in this regard, as it builds on the direct cooperation with \lidar companies to offer better integrations.
However, while there exist relatively realistic physics based simulations, these are locked into the Nvidia ecosystem, so we see them as rather complementary to our accessible data based approach.

Furthermore, our approach makes it possible to learn the behavior of new LiDAR sensor boxes only based on test drive data, without knowing or tuning the internal physical parameters.

\paragraph{Data Sets}
Many high quality data sets from test drives have emerged in recent years \cite{sun_scalability_2020}
\cite{Uhrig2017THREEDV}
and the trend continues~\cite{Liao2021ARXIV}.
\lidar data is very common to be included in these data sets, which makes them compatible with our approach.
The usual representation format are \lidar point clouds. 
For each such point, there is a position, an intensity value and a time stamp.
Using the sensor orientation and offset between the RGB camera, it is possible to compute a 2D \lidar projection.
Often, this format is already included in the data sets directly or the necessary projection matrices are given.
Notably, \cite{gaidon2016virtual,cabon2020vkitti2} replicates \kitti scenes virtually with a semi-manual process, thereby providing ground truth data like depth and semantic segmentation.
%For our pipeline, however, we require real world \lidar data so the network can learn the sensor behaviour so we can not leverage this data set for training. 
%However, 
This allows us to compare how well our trained network is able to generalize between synthetic and real data on the same driving data sequences.
%\MS{is one of these a2d2? Should we mention here, which data sets we use?}

\paragraph{Image-to-image Translation}
In a way, the 2D \lidar projection can be interpreted as a special segmentation of the RGB image using two classes: visible by the \lidar sensor and not visible. 
Consequently, the popular U-Net architecture~\cite{ronneberger2015unet}, using an encoder network, followed by a decoder network, immediately suggests itself.
However, to simulate the actual \lidar pattern in the projection, it is also important that unrealistic outputs are penalized. 
This can be achieved efficiently with General Adversarial Networks (GANs)~\cite{goodfellow2014generative}, where a discriminator is optimized that estimates whether an image is fake or not.
Pix2pix~\cite{isola_image--image_2018} combines the strength of U-Nets and an adversarial loss and demonstrated success for very diverse image domains.
There are also approaches that perform this task in an unpaired fashion~\cite{zhu_unpaired_2020}, but for the pipeline proposed in this paper, we can assume that paired images exist.

Building on pix2pix, pix2pixHD~\cite{wang2018pix2pixHD} and  SPADE~\cite{park2019SPADE} have improved the quality of results via adapting the architecture more towards specific data domains.%, e.g., using the semantic segmentation to learn predictions in relation to the labels.
An alternative would be the direction of style transfer~\cite{johnson_perceptual_2016}.
Following this approach, one could use an average \lidar projection and apply it to different images.
Given the available data, which offers a specific \lidar projection for each individual input image, image-to-image translation is a better fit for our purposes.

\paragraph{Learning based Sensor Simulation}
A very interesting approach for \lidar simulation is pseudo-\lidar~\cite{wang_pseudo-lidar_2018}.
It is similar in regard to how point clouds are generated based on RGB images and depth prediction networks. 
However, the end goal of this method is to improve image based object detection, while our pipeline is more general and generates  realistic \lidar data on purpose as an intermediate representation to be used to train various downstream applications.

On a conceptual level, we pursue the same approach as L2R GAN~\cite{Wang_2020_ACCV}, but try to learn the mapping between camera images and camera \lidar projections instead of the one between bird's eye view \lidar and radar images.

The closest approach to ours is LiDARsim~\cite{manivasagam_lidarsim_2020}.
It also aims at  realistic simulation of \lidar leveraging real world data but use a physics based light simulation as foundation.
To obtain realistic geometry, a virtual copy is reconstructed by merging multiple scans from test drives. %, which allows  appropriate features can be extracted for the neural network.
Instead of directly predicting the \lidar point distribution based on camera images, LiDARsim first uses ray casting in this virtual scene to create a perfect point cloud.
Based on these and the \lidar scans from the test data, the network then learns to drop rays from the ray casted point clouds for enhanced realism.

With our method, we try to achieve a more lightweight \lidar simulation for situations where such preprocessing is not feasible. 
For direct integration into external simulation environments, a solution is required that does not dependend on high quality geometry data.
%An alternative to this could be data that is readily available like RGB images and more abstract like semantic segmentation masks.

\section{\uppercase{Training and Simulation}}

\subsection{\lidar Images}
To learn and simulate the behaviour of \lidar sensors, we map our problem to an image-to-image translation task.
Therefore, we convert the \lidar output to an image which we call {\em \lidar image}.
We notice that the problem of \lidar simulation in a virtual environment is not the measured distance of a particular sample, as this can easily be determined by ray casting.
Instead, we need to decide {\em which} \lidar rays return an answer and which do not, e.g.~due to absorption, specular reflection, or diffusion.
We map this information to a \lidar image, that tells us for a particular view direction, whether a ray in this direction is likely to return an answer.

\begin{figure}
	\centering

	\begin{subfigure}{0.49\textwidth}
		\centering
		{\includegraphics[width=1\textwidth]{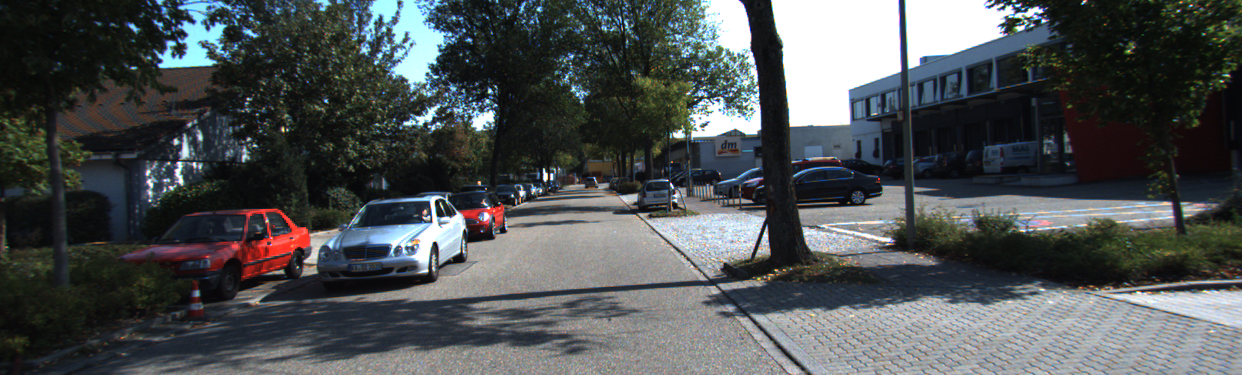}}
		\caption{RGB image from the \kitti data set.}
	\end{subfigure}

	\begin{subfigure}{0.49\textwidth}
		\centering
		{\includegraphics[width=1\textwidth]{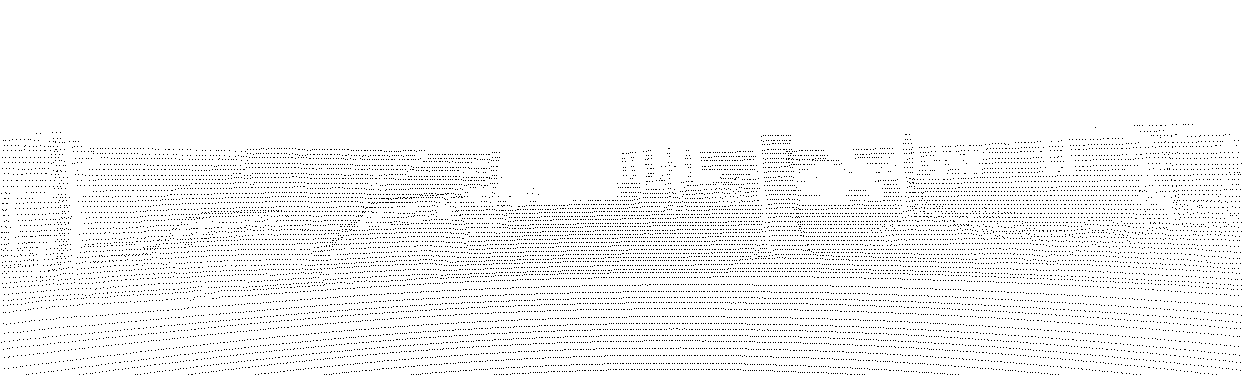}}
		\caption{\lidar points projected into same view.}
	\end{subfigure}
	
	\begin{subfigure}[b]{0.49\textwidth}
		\vspace*{5mm}
        \begin{tikzpicture}[zoomboxarray,
    		zoomboxes below,
    		zoomboxarray columns=3,
    		zoomboxarray rows=1,
    		connect zoomboxes,
    		zoombox paths/.append style={ultra thick, red}]
    		\node [image node] { \includegraphics[width=\linewidth]{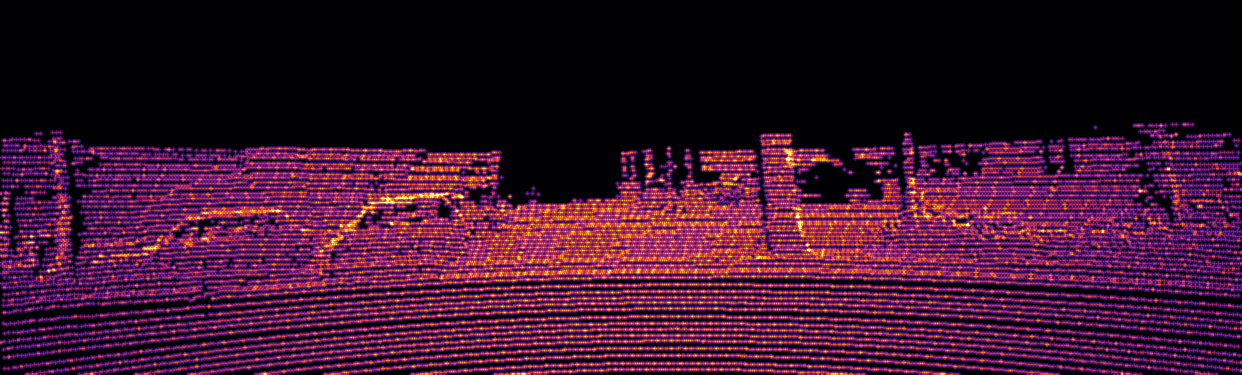} };
    		\zoombox[magnification=2]{0.335,0.42}
    		\zoombox[magnification=6]{0.335,0.42}
    		\zoombox[magnification=40]{0.881,0.660}
    	\end{tikzpicture}
    	\caption{Blurred and color coded \lidar image.}
    \end{subfigure}
	\begin{subfigure}[b]{0.49\textwidth}
		\vspace*{5mm}
    	\includegraphics[trim={5cm 0 9,8cm 0},clip,width=0.49\linewidth]{pictures/heat/1_0000000059_tar_heat.png}
    	\includegraphics[trim={0 0 0 11,7cm},clip,width=0.49\linewidth]{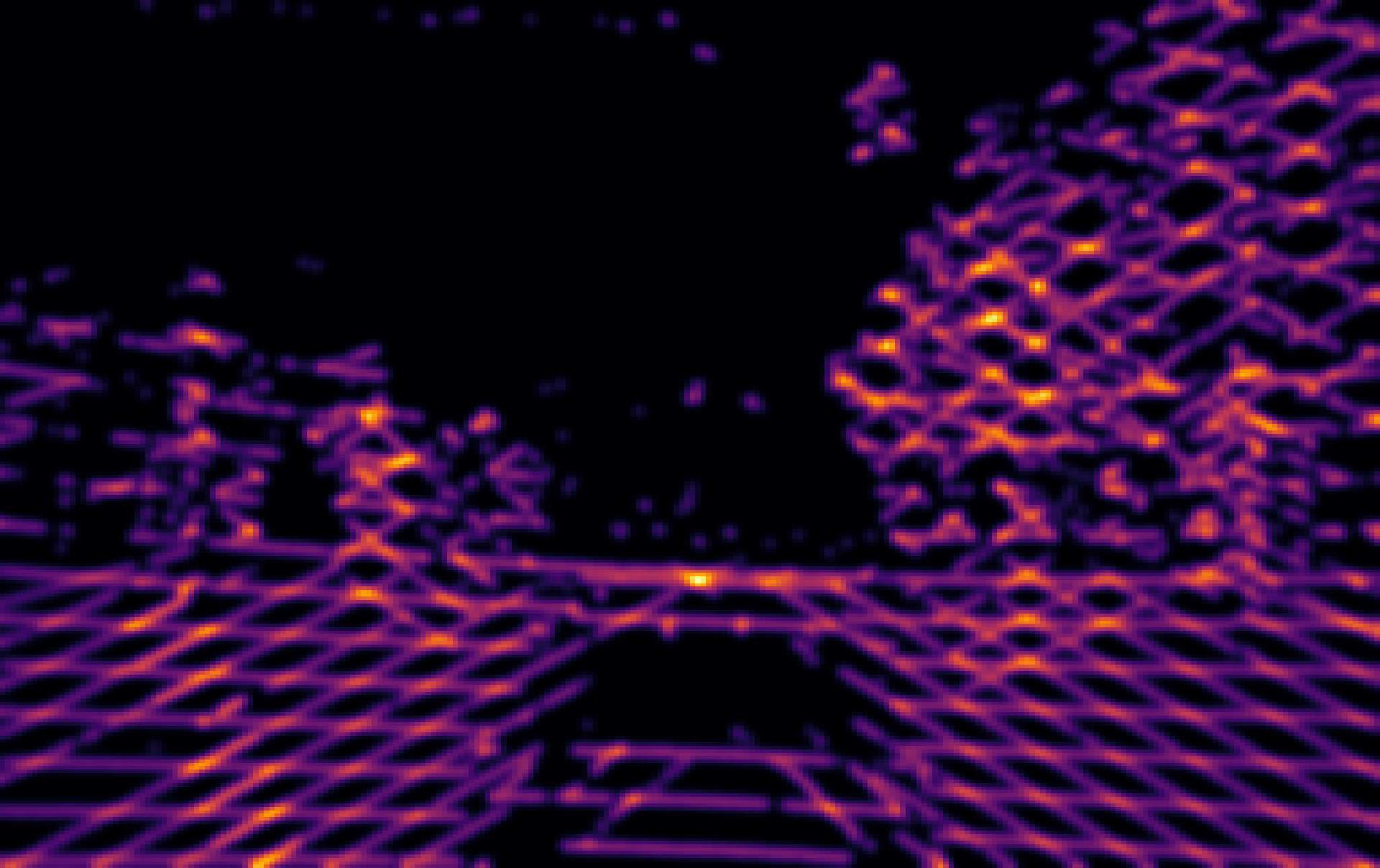}
	    \caption{\lidar images from two different data sets (left: \kitti, right: \di with five sensors resulting in a grid pattern)}
	\end{subfigure}
	\vspace*{1mm}
	\caption{\lidar image generation }
	\label{fig:lidarimage}

\end{figure}

To this end, we project the \lidar point cloud into the camera view, which requires that we know the relative pose of \lidar and camera and that the outputs of all devices are aligned temporally and have a timestamp.
An example is shown in Fig.~\ref{fig:lidarimage}(b).
The binary image shows, which regions are visible for the sensor, but also which problems arise when trying to predict them:
Since camera and \lidar sensor usually do not have the same optical center,
%\footnote{As a noticeable exception, latest \lidar sensors can also act as a camera with then identical optical center.} 
the position of the projection varies with depth.
A network has to predict depth implicitly to output dots at the proper position.
%Second, binary images lack a gradient that can be used in learning, so image translation networks are better in predicting smoother, non-binary images.

We thus blur the binary image with a simple Gaussian filter, as shown in Fig.~\ref{fig:lidarimage}(c),
Note that we color coded the result for better visualization.
The resulting map is denser, so (i) it better abstracts from the exact sample positions and (ii) the map is easier to predict.
The size of the Gaussian filter is chosen such that gaps between neighboring dots are filled, but also such that invisible regions are not filled up wrongly.
It depends on the scan pattern of the \lidar and the image resolution, which we choose in turn depending on the horizontal and vertical distribution.
In our examples, we use a Gaussian blur with $\sigma = 8$ to make up for the sparsity in the \di data set (image resolution $1920\times 1208$) and a rather small intervention for \kitti (image resolution $1242 \times 375$), only using a small 5x5 custom kernel as shown in Fig.~\ref{fig:lidarimage}(c). 
%\MS{image resolutions? I guess blurring happens before scaling to 256x256?}
There are also more elaborate ways of augmenting the data~\cite{shorten_survey_2019}, but the basic blurring we use is well suited for \lidar data and resulted in a sufficient improvement of the training results.

The blurring also helps to hide another, more subtle inconsistency: usually, the camera frame rate is higher than that of the \lidar.
However, to get a full \lidar pattern, we project all points from a single complete \lidar scan to each camera frame.
Additionally, each \lidar sample comes from another time interval, resulting in a rolling shutter effect.
By blurring and interpolating all this, we achieve a less sharp, but more consistent visibility mask.
%
% Note that blurring the images requires different interpretation during later steps of the pipeline. 
% Instead of a direct point distribution, we arrive at a visibility map, which can then in turn be used to sample the individual \lidar points according to the visibility values (see Sec.~\ref{sec:integration}.
%
The \lidar image also contains information about the scan pattern of the \lidar, and can even be used to represent the result of multiple \lidar sensors, as in the A2D2 dataset shown in Fig.~\ref{fig:lidarimage}(d).

%There are also existing datasets that go beyond that and align the point clouds with the camera images in a post-process, but this is not crucial for our pipeline.

%\begin{figure}[!h]%
%	\centering
%	{\includegraphics[width=0.48\textwidth]{pictures/hits/1_0000000059.png}
%		\caption{make these brighter}}
%	\label{fig:1}
%\end{figure}

% An important step is to increase the density of the \lidar projections. 
% In comparison to the input image, the \lidar resolution is rather low (cf. Figure~\ref{fig:kitti_pair}), which inhibits the learning process of the neural network as large areas of the images carry no information.
%Using a resolution of 256x256 already helps when using appropriate downscaling.
%One possible solution here is to downscale the camera image first and then reproject the point cloud into the lower resolution image.
%A more streamlined solution for our purpose, however, 
% A robust solution for this problem is to blur the high resolution \lidar projections before downscaling. 
%This is roughly equivalent but also gives us more control over the images.

\subsection{\lidar Simulation}

Our hypothesis is that an image-to-image translation of an RGB image to a \lidar image should be well possible, since in an RGB image it is possible to detect \lidar-critical regions, such as sky, far distances, or transparent/mirroring surfaces (e.g., windows).
However, it is unclear how well the network trained on real data generalizes to rendered and thus much cleaner images.

We furthermore hypothesize that a translation to \lidar images is also possible from depth images or from segmentation images.
Such images are very simple to generate in a driving simulation, however these are not generally available as ground truth data for training and thus have to be derived (e.g.~using depth estimation networks), or generated manually (e.g.~segmentation masks).

\begin{figure*}[ht]
	\centering
	\includegraphics[width=\linewidth]{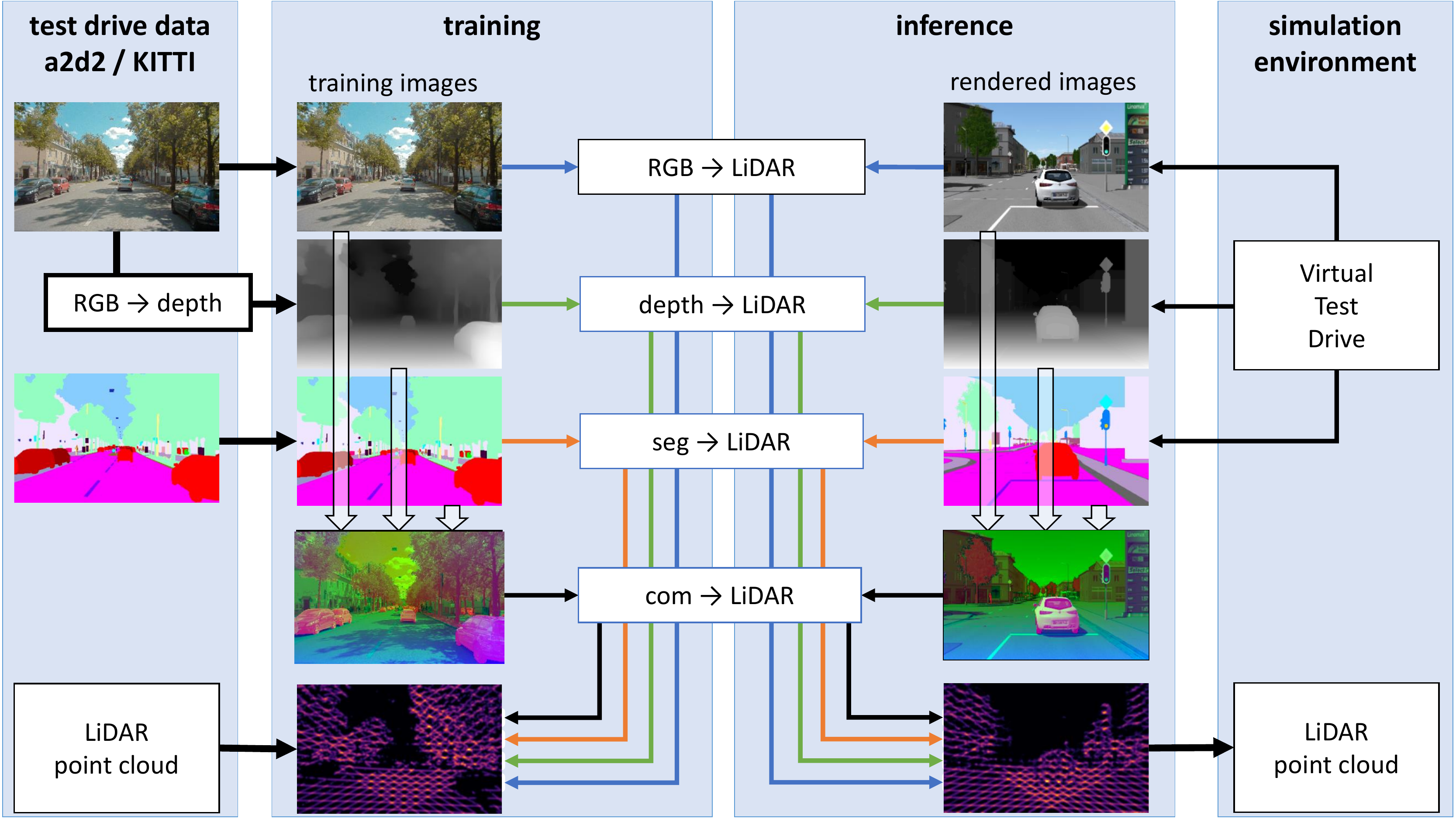}
	\caption{Overview over training and inference of our \lidar simulation. We explore prediction of \lidar point clouds based on RGB, depth and segmentation images separately and combined. We train on real RGB and \lidar pairs and apply the learned mapping on rendered image to synthesize realistic \lidar images.}
	\label{fig:system}
\end{figure*}

In the following, we explore all these options, see Fig.~\ref{fig:system}.
In particular, we describe how to generate the input data, how our image-to-image transformation is done, and how we integrated all this into a virtual test environment.

\subsection{Training Data}
\label{sec:input}

We use two data sets to generate our training data:
\begin{itemize}
	\item \kitti: 
	%We use the split of training and test data as given by \cite{uhrig2017sparsity}, resulting in 42.949 training images and 3.426 test images.
	We use the training set~\cite{Weber2021NEURIPSDATA} from the Segmenting and Tracking Every Pixel Evaluation benchmark, since it also  includes segmentation data. We make a slight variation to the validation set and only include the sequences that are present in VKITTI. This results in roughly 6000 training and 2000 validation images

	\item \di:  We leave about 20000 images for training and create a validation set from three different test drives, one in a urban setting, one on a highway and one on a country road making up 6000 images.
	In contrast to many other data sets, \di uses five low resolution \lidar sensors in tandem to achieve sufficient point density.

\end{itemize}

The \kitti and \di dataset create different \lidar patterns, which allows us to test our approach on different point layouts (see Fig.~\ref{fig:lidarimage}d). Our learning based approach overcomes the need to implement the physics of each sensor for implementation and enables to learn the sensor properties for a simulated environment.

As input to our image-to-image network we consider RGB camera images, depth images, and/or segmentation masks.
Ground truth depth maps are usually not available, so we derive them using mono depth estimation \cite{Ranftl2021}.
%},  which is combined with a Transformer~\cite{vaswani2017attention} architecture~\cite{
If ground truth segmentation masks are not available, they can be estimated using segmentation networks, e.g.~\cite{Ranftl2021,pointrend}.
As forth option, we combine camera, depth, and segmentation images by merging a grey scale version of the camera image, depth, and segmentation label.
To avoid the use of the same value for different semantic objects, the colors of the class labels are mapped to specific grey values.
Overall, we follow the set of more than 50 different classes defined by \di.
The color encoding also includes instance segmentation of up to four instances per class so that overlapping objects can be discerned.

\subsection{Image-to-image Translation}

Many different network architectures for image-to-image translation have been presented, our experiments have shown that pix2pix and even simpler encoder-decoder architectures like U-Nets \cite{ronneberger2015unet} deliver satisfying results as also shown by LiDARsim~\cite{manivasagam_lidarsim_2020}. % and in our Appendix Figure~\ref{fig:unet}.
%Advanced networks that build on pix2pix like pix2pixHD or SPADE could potentially improve results but are less flexible regarding the image format \BE{how?}, so we limited our experiments to the original pix2pix architecture.
We apply a widely used implementation\footnote{\url{https://tensorflow.org/tutorials/generative/pix2pix}} which is based on a U-Net architecture and a convolutional PatchGAN classifier as proposed in the original pix2pix paper \cite{isola_image--image_2018}.
We leave the network architecture unchanged but omit the preprocessing step that crops and flips the pictures randomly.
This step is problematic for \lidar projections as they are dependent on the camera perspective and not necessarily symmetric.
% Since the data sets we use intrinsically already show similar images with slight variations between subsequent frames, the general idea of this step still applies implicitly. \MS{?}

%Training pix2pix networks is completed relatively fast, typical training times are in the range of \MS{todo}.
% as a local optimum reproduces the overall \lidar pattern and the most prominent interferences \MS{?} in the 2D projection.
Our experiments have shown that deviations in the basic parameters of the architecture only show little effect.
For the presented results we used the following configuration: Adam optimizer with learning rate of $2 * 10^{-4}$ and $\beta_1 = 0.5$ and $\lambda = 100$ for computing $ \text{Generator Loss} = \text{Adversarial Loss} + \lambda * \text{L1 Loss}$. 

\subsection{Integration}
\label{sec:integration}
Once training is completed, we integrate the learned \lidar sensor network into our simulation environment and 
%Depending on which image-2image translation variant we choose (RGB, depth, segmentation, or combined), we generate the required input images directly on the GPU, as shown in Fig.~\ref{fig:system}.
% The target format can vary from the individual points of the \lidar projection to full visibility maps.
our network then predicts a \lidar image, as shown in Fig.~\ref{fig:system}. %which we finally convert back to a \lidar point cloud.
To convert back to a \lidar point cloud, a simple approach is to sample the map directly:
\begin{enumerate*}[label=(\roman*)]
	\item sample the visibility map, e.g.~on a uniform grid,
	\item discard all points with small visibility,
	\item determine the remaining samples' depth from the depth map,
	\item reproject the points into a 3D coordinate space.
\end{enumerate*}

The resulting point cloud mimics the behavior of the real sensor, in particular typically invisible points are missing, but the original \lidar pattern is probably not represented well.

A more expensive, but also more precise solution is to generate virtual rays for the original \lidar samples, compute their hit point by ray casting, and check the visibility of the hit point in the predicted \lidar visibility map.
This approach can also consider the car movement during a \lidar scan and thus reproduce the rolling shutter effect.

\section{\uppercase{Experiments and Results}}\label{sec:results}

Overall, our implementation achieves a visually reasonable mapping between input image and \lidar projection.
The training on the data translates well to the test data sets and also generalizes on synthetic data to a sufficient degree.

%We show one example in Figure~\ref{fig:kitti_example} but more can be found in the Appendix\BE{can you point here to the exact figures in the appendix?}.

%In the following, we describe experiments with different data configurations and evaluate the results that can be achieved with our pipeline.

%Aside from RGB camera images, we also use depth and segmentation masks as training input.
%For the latter, we use the annotated images from \di.
%As there are no complete depth maps for neither \di nor \kitti, we generated them ourselves via monocular depth estimation. 
\begin{figure*}[!h]
	\centering
	\begin{subfigure}[b]{0.225\textwidth}
		\centering
		{\includegraphics[width=1\textwidth]{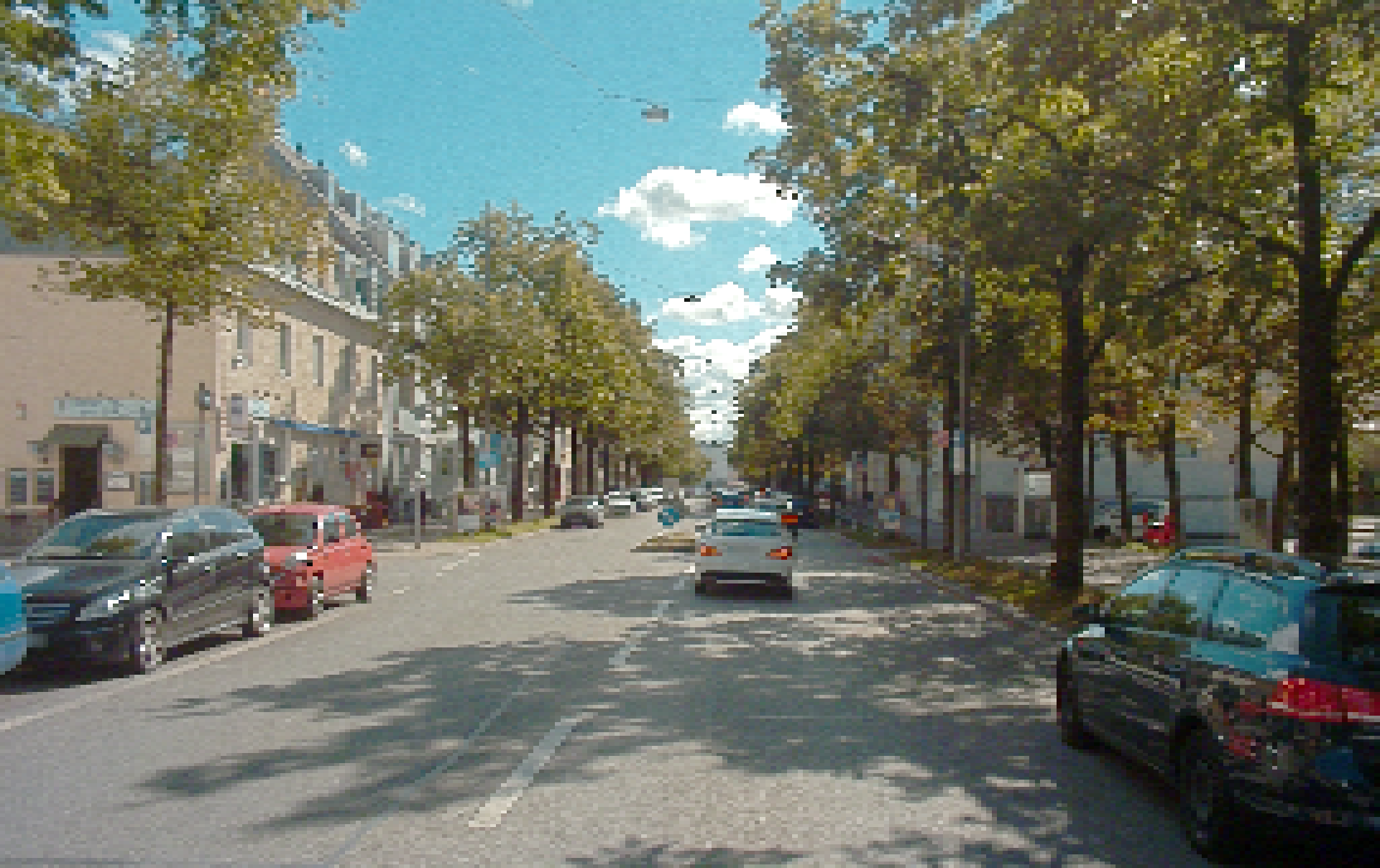}}
	\end{subfigure}
	\hfill
	\begin{subfigure}[b]{0.225\textwidth}
		\centering
		{\includegraphics[width=1\textwidth]{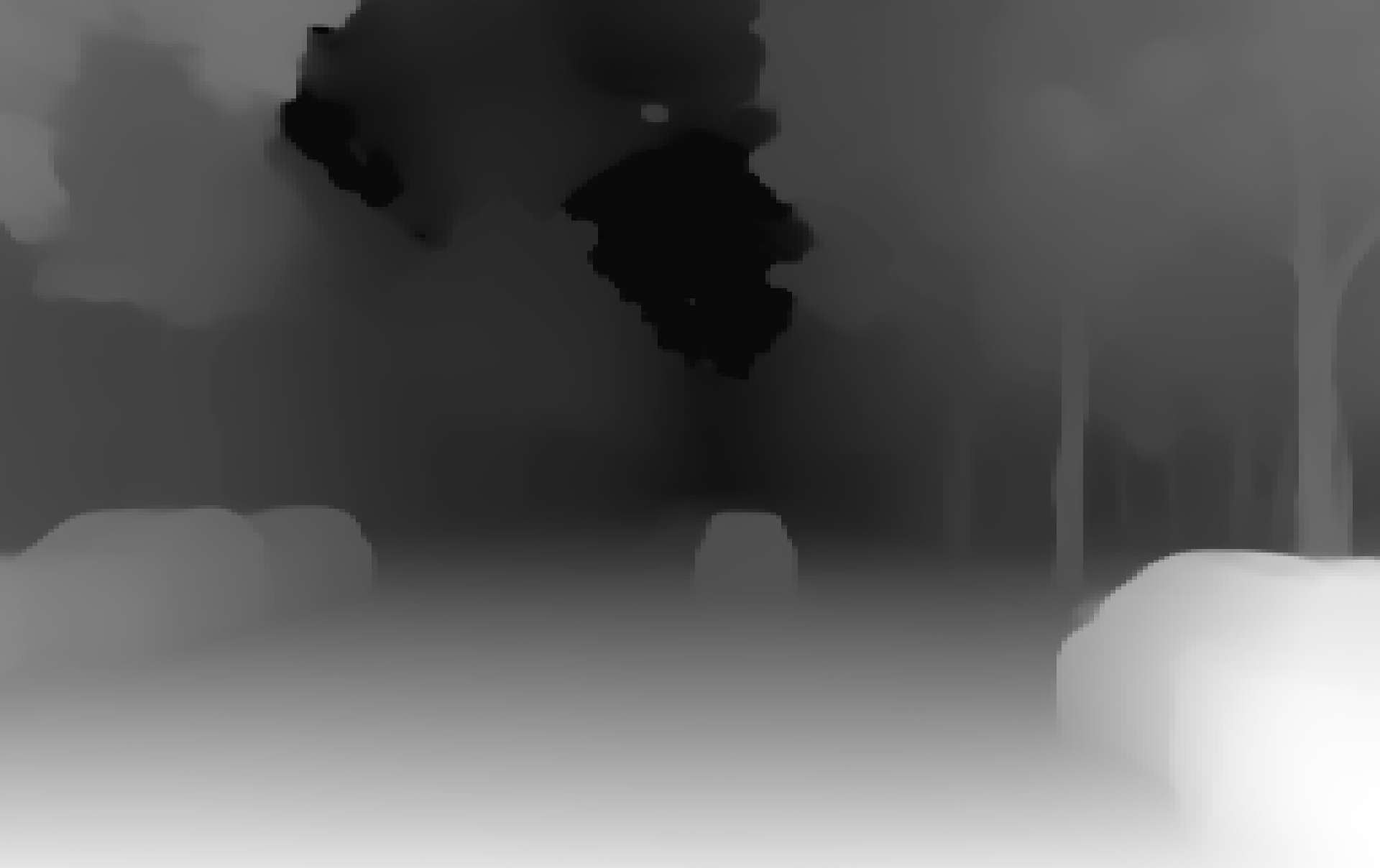}}
	\end{subfigure}
	\hfill
	\begin{subfigure}[b]{0.225\textwidth}
		\centering
		{\includegraphics[width=1\textwidth]{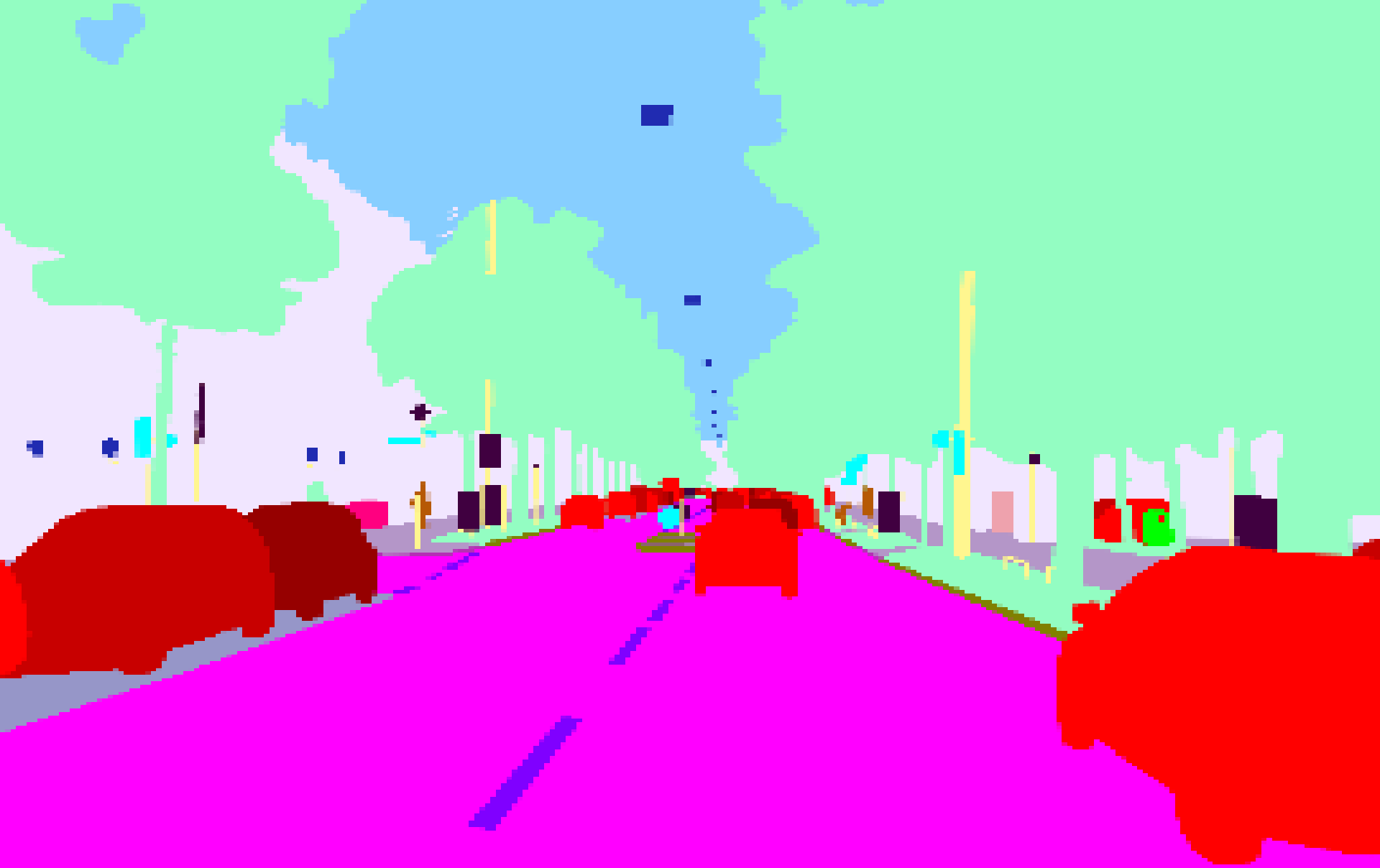}}
	\end{subfigure}
	\hfill
	\begin{subfigure}[b]{0.225\textwidth}
		\centering
		{\includegraphics[width=1\textwidth]{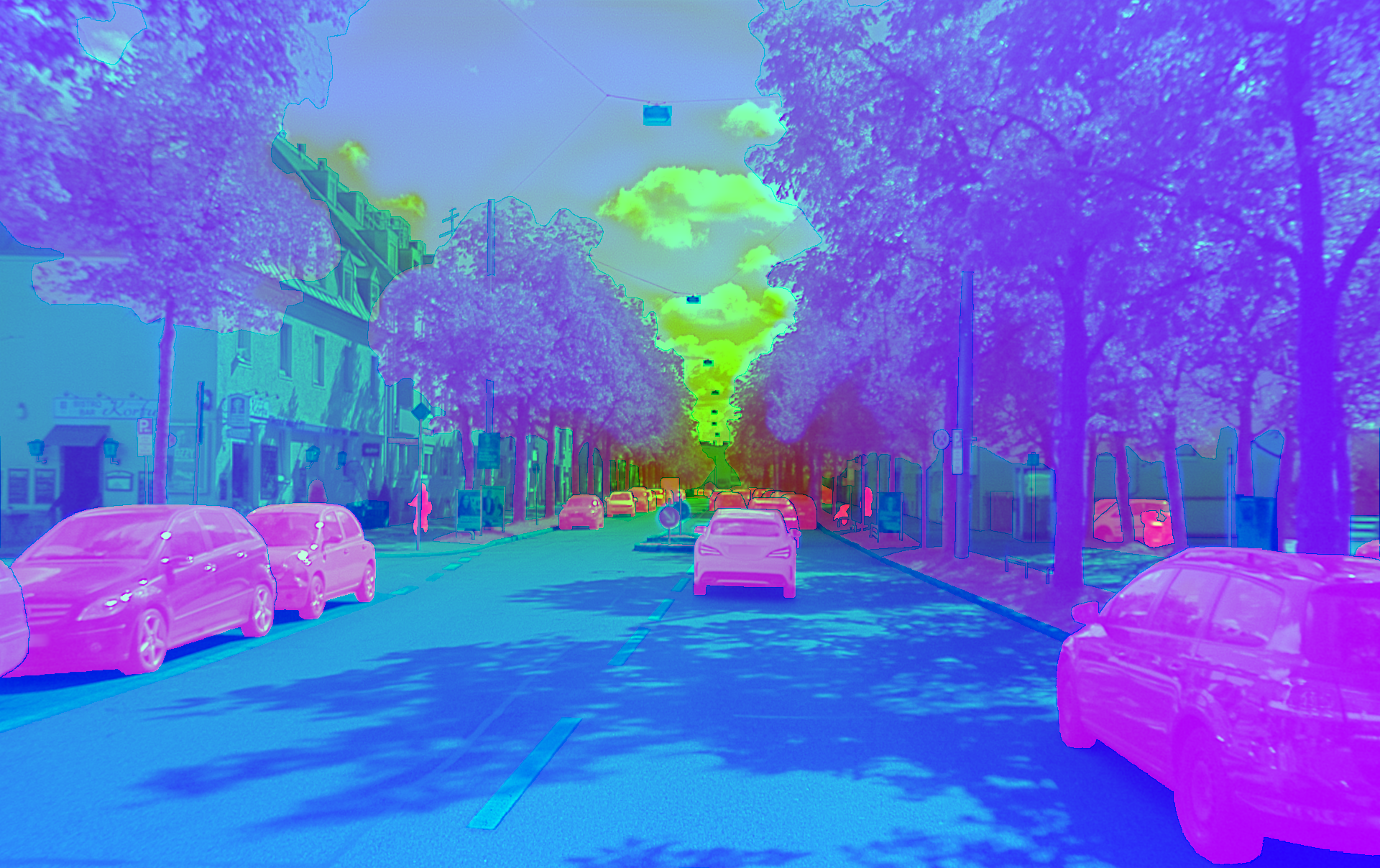}}
	\end{subfigure}
	
	\begin{subfigure}[b]{0.225\textwidth}
		\centering
		{\includegraphics[width=1\textwidth]{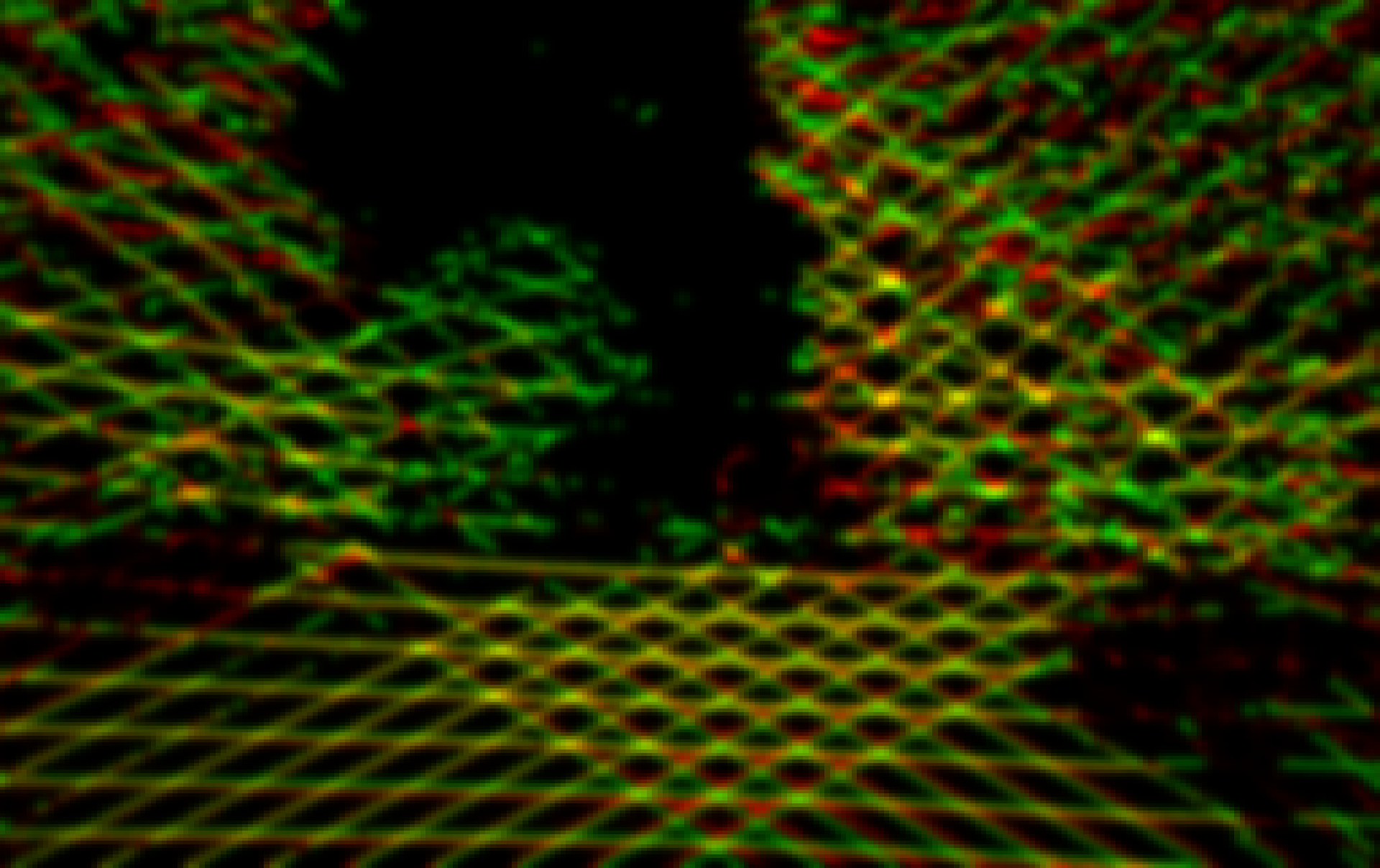}}
		\caption{Trained on Camera Images}
	\end{subfigure}
	\hfill
	\begin{subfigure}[b]{0.225\textwidth}
		\centering
		{\includegraphics[width=1\textwidth]{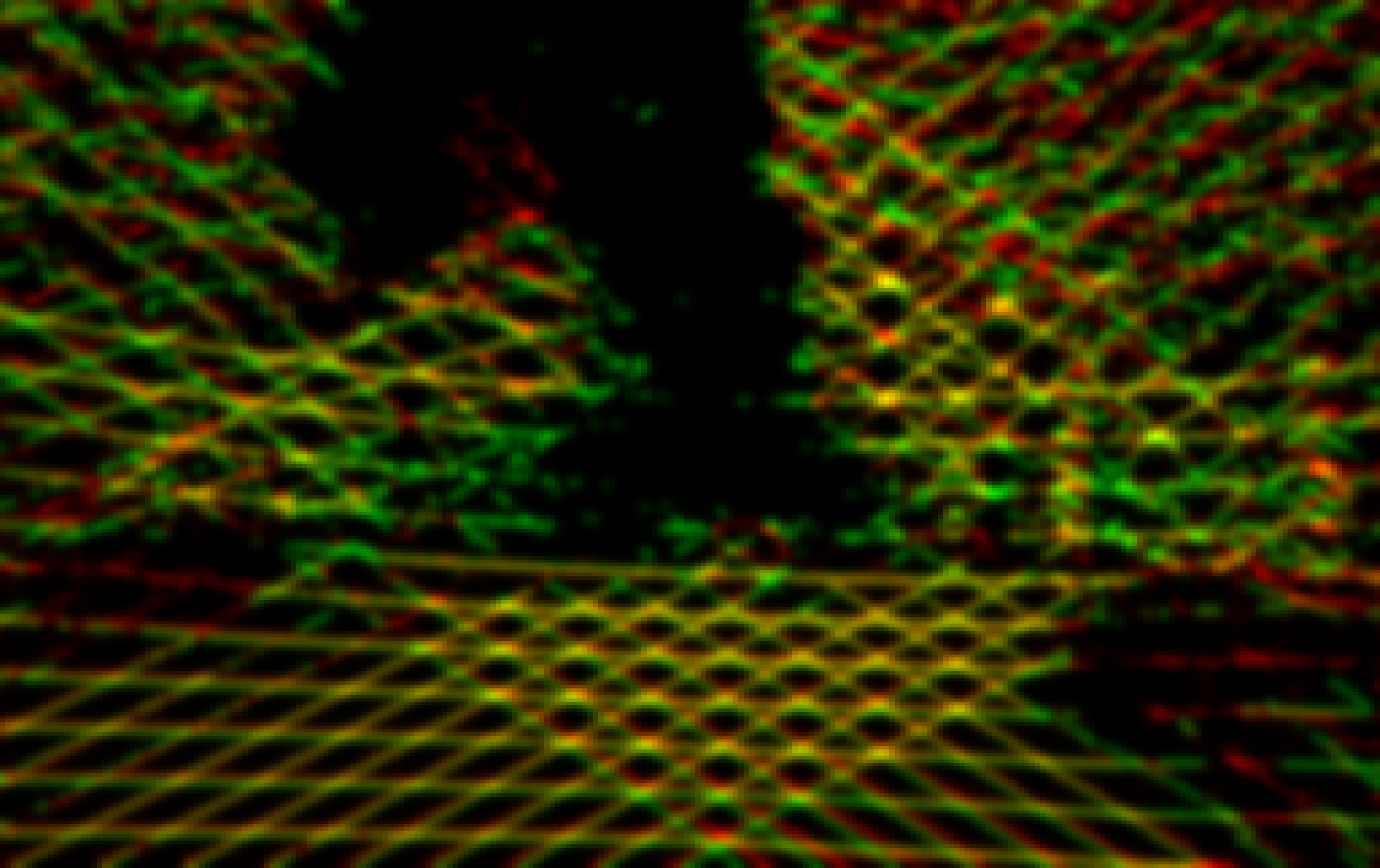}}
		\caption{Trained on Depth Maps \hspace{20 mm} }
	\end{subfigure}
	\hfill
	\begin{subfigure}[b]{0.225\textwidth}
		\centering
		{\includegraphics[width=1\textwidth]{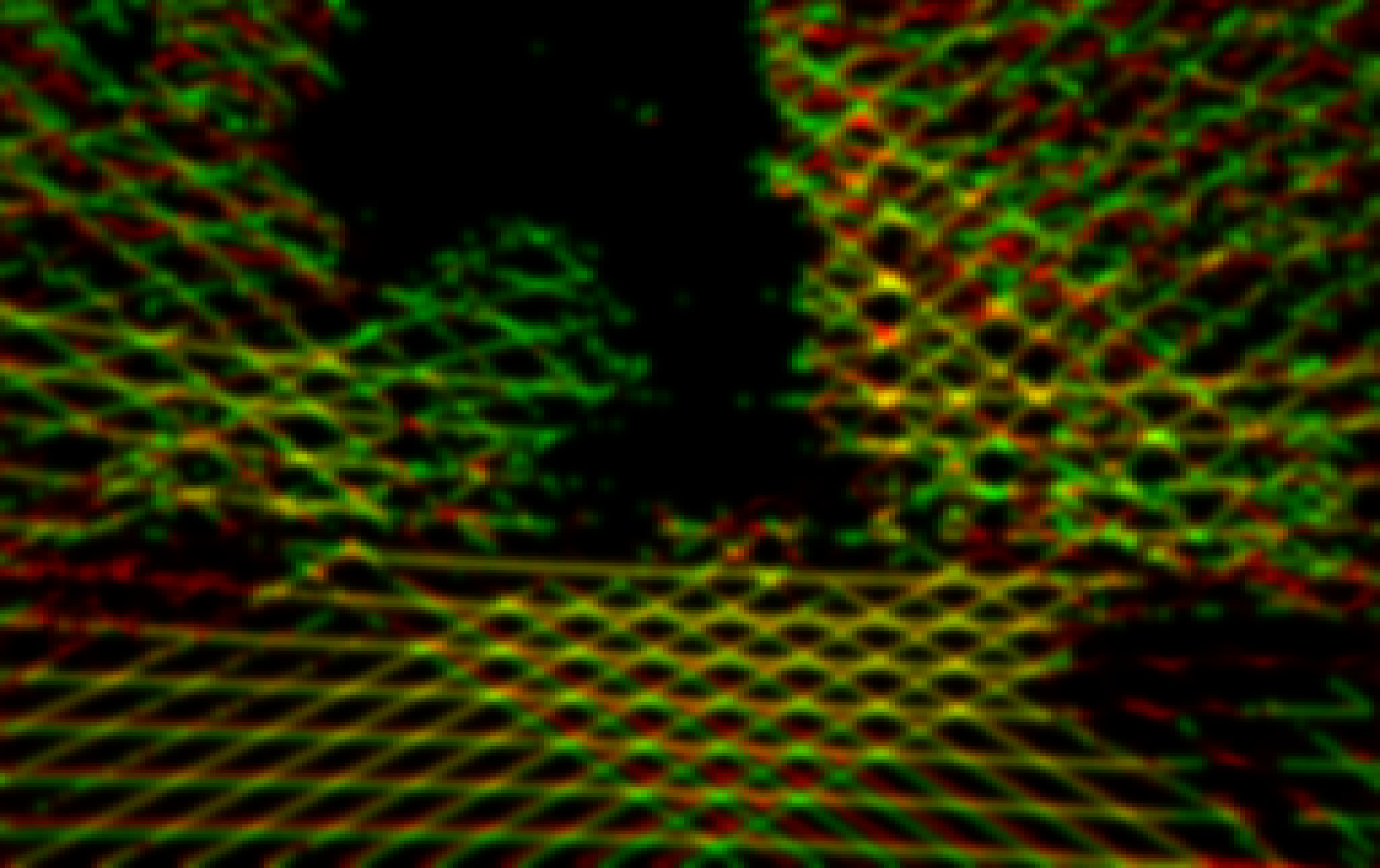}}
		\caption{Trained on Segmentation Maps}
	\end{subfigure}
	\hfill
	\begin{subfigure}[b]{0.225\textwidth}
		\centering
		{\includegraphics[width=1\textwidth]{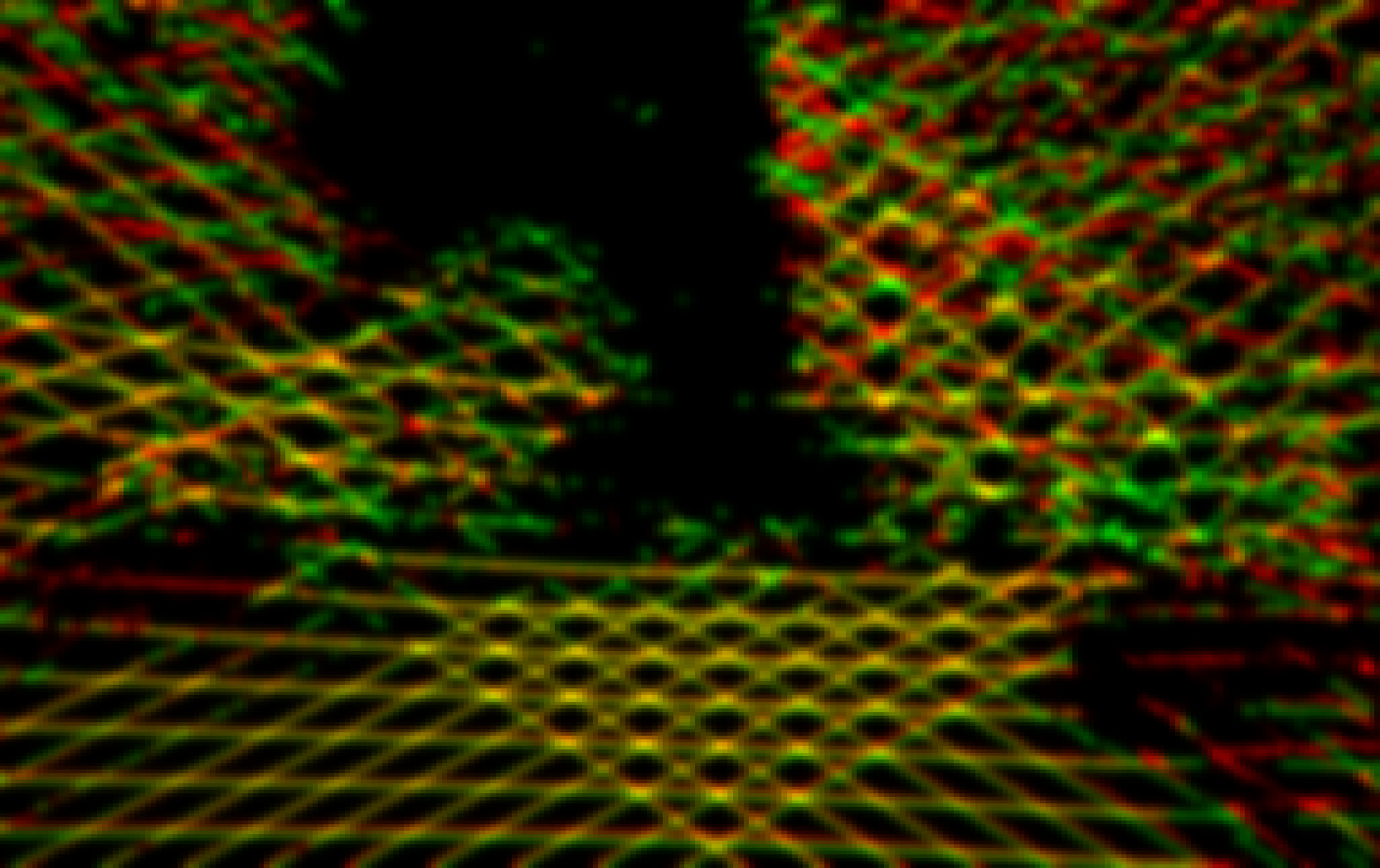}}
		\caption{Trained on Combined Images}
		
	\end{subfigure}
	
	\vspace{+0.1cm}
	\caption{Results on the \di data set: lower image shows prediction (red) and ground truth (green), correctly predicted pixels are consequently yellow.}
	
	\label{fig:results_a2d2}
\end{figure*}

\subsection{Qualitative Evaluation}
Both, \kitti and \di work with the same network configuration and only require different filter kernels to be applied as described in Section~\ref{sec:input}.
Naturally, there are significant differences resulting from the different training data sets and sensor configurations; \kitti has higher \lidar density, but a smaller upwards angle.
This means that elevated objects like traffic lights or trees are often not detected correctly.

Due to the reduced number of training samples compared to \di, the \kitti results lack in visual quality.
%Vice versa, this means that the approach should scale very well with larger data sets.
Concerning the sensor artifacts, we observe that the general projection pattern is acquired very well by the network. 
The points in the distance are also cut as limited by the sensor range but especially the variants without depth are less effective here.
Individual objects, however, are harder to distinguish by eye and reflection effects are inconsistent.
With a learning based approach, there is an upper limit to what is possible, given the available input.
In reality, there are objects that appear very similar on camera but have very different reflection attributes. %and reflective surfaces are comparatively rare.

\subsection{Quantitative Evaluation}

For training we have used an Nvidia RTX 3080.
We use 32k training steps
% for every configuration,
which can be completed in less than half an hour, so our pipeline can be trained quickly for new sensor configurations.
Depending on the data set, the network has not necessarily seen every input picture once, but since consecutive frames are very similar, more training has shown no benefit in our experiments.
Inference on the same GPU takes 17ms without any optimizations, which allows real-time \lidar simulation. %which is less than the time a real \lidar sensor requires for one scan.
%Ignoring other tasks, this allows the simulation to run at almost 60 FPS, which suits real-time requirements for many applications. 

%\subsubsection{Performance on Validation Data}
To evaluate the quality of our results, we predict \lidar images on real-world input from \kitti and \di and compare these with ground truth \lidar data.
Depth images are estimated from the RGB images and as segmentation masks we use ground truth data from the data sets.

\begin{table}[h]
	%\vspace{-0.2cm}
	\caption{Evaluation on \kitti , \di and VKITTI: Errors on respective test set in percent.}\label{tab:example1} \centering
	\begin{tabular}{|c|c|c|c|c|}
		\hline
		& $L1$  		& $L1^+$   &  $L1^-$  & $L2$\\
		\hline
		\hline
		\tiny\kitti RGB	    & 8.64 & 6.14  & 2.50 & 14.33 \\
		\hline
		\tiny\kitti Depth	&8.08 & 4.92  & 3.16  & 13.58\\
		\hline
		\tiny\kitti Sem	    & 8.72 & 5.90  & 2.82  & 14.44\\
		\hline
		\tiny\kitti Com	    & 8.63  & 4.96  & 3.67 & 14.36\\
		\hline
		\hline
		\tiny \di RGB	& 10.52 & 5.44  & 5.08 & 17.02 \\
		\hline
		\tiny \di Depth & 10.22 & 5.36  & 4.86  & 16.63\\
		\hline
	    \tiny \di  Sem	& 10.55 & 10.16  & 0.39  & 17.02\\
		\hline
		\tiny \di  Com	& 10.38 & 6.01  & 4.38 & 16.90 \\
		\hline
		\hline
		\tiny V\kitti RGB	& 8.33 & 4.28  & 4.05 & 14.06 \\
		\hline
		\tiny V\kitti Depth	& 8.64 & 4.88  & 3.76  & 14.26\\
		\hline
	    \tiny V\kitti Sem	& 8.26 & 3.94  & 4.32  & 13.89\\
		\hline
		\tiny V\kitti Com	& 8.39 & 4.71  & 3.68 & 13.98 \\
		\hline
	\end{tabular}
\end{table}
We compute the $L1$ error ($L1=|A-B|$) where $A$ is the prediction and $B$ is the ground truth). %that also has been used during training on the test sets.
%\BE{you never said that you used L1 loss in your pix2pix implementation and I think it uses a GAN loss, no?:)}
Moreover, we define a positive error $L1^+=\max(A-B,0)$, where the network predicts points that are not present in the ground truth, and a negative error $L1^-=\max(B-A,0)$, where points are missing in the prediction.
Additionally, we compute the $L2$ error.

The results in Tab.~\ref{tab:example1} show that the prediction results for RGB, depth, segmentation, and combined are relatively close.
Prediction on \kitti data works better than for \di, which partly results from the larger black space in the \lidar projections.
Depth seems to work generally well, and since it is generated in any simulation anyhow, it looks like a good candidate for integration.
However, one would have to examine closer, whether features important for the following task are well represented, e.g.~car windows.
The combined mode showed no advantage, however smarter ways to combine the three modalities might deliver better results.

Overall, the quality of images trained in a network like pix2pix cannot fully be estimated by pixel-wise errors.
Small shifts in the \lidar pattern, for example, result in large errors.
%If the network returns an almost correct image that is simply shifted by a few pixels, the error would be large but the result would be a good simulation.
Fig.~\ref{fig:results_a2d2} overlays the ground truth and the prediction, showing deviations between the different input types, most prominently in the distant parts, where a correct depth estimation becomes important.

%Analysing the different error values, it needs to be clear that the error moves in a rather low percentage by default because big parts of the target picture are simply black.
%By increasing the point density, however, we have counteracted this to some degree.
% \RM{die Daten hab ich jetzt nicht ganz geschafft zu interpretieren, ich hab aber noch einen Bug gefixt, die Prozentwerte sind jetzt wieder niedrig, siehe txtfile, Marc, vielleicht kannst du da noch was schreiben, du hattest ja auch die Idee, auf die Unterschiede zwischen RGB und Rest zu fokussieren, ich denke die Daten geben das auch tatsächlich wieder}

%As the values are generally very close to each other throughout all experiments, the most interesting outlier is the large negative error compared to the positive error in the semantic segmentation map results.
%This indeed seems logical as the segmentation maps have the least depth information and thus can lead to problems when deciding whether to draw points in the distance or not.

\begin{figure}[!h]
	\centering
	\begin{subfigure}[b]{\linewidth}
		\centering
		{\includegraphics[width=\textwidth]{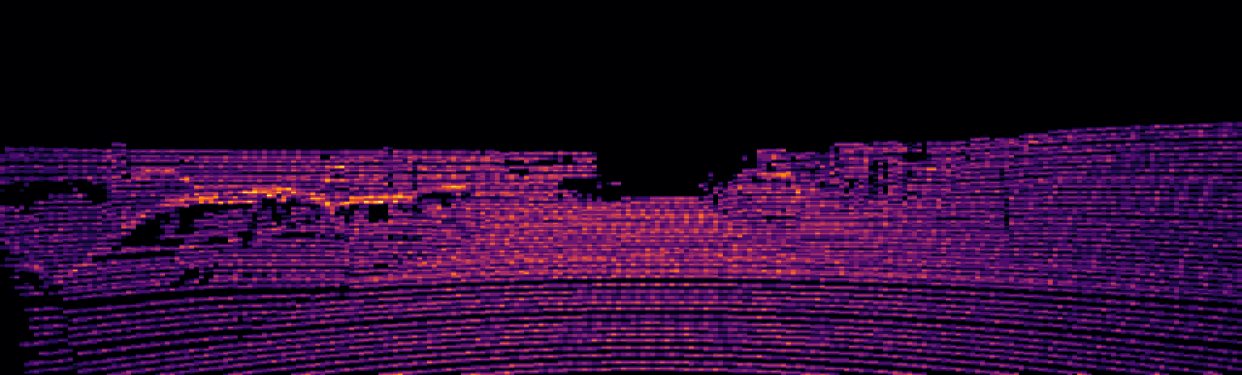}}
		\caption{Ground Truth \lidar Image}
			\vspace{1em}
	\end{subfigure}
	%\hfill

	\begin{subfigure}[b]{\linewidth}
		\centering
		{\includegraphics[width=\textwidth]{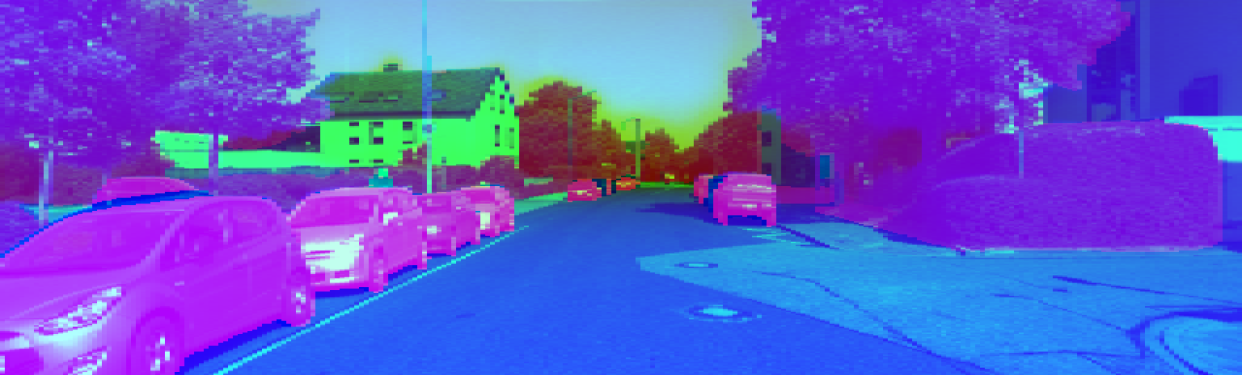}}
		%\caption{Combined \kitti Input Image}
	\end{subfigure}
		\begin{subfigure}[b]{\linewidth}
		\centering
		{\includegraphics[width=\textwidth]{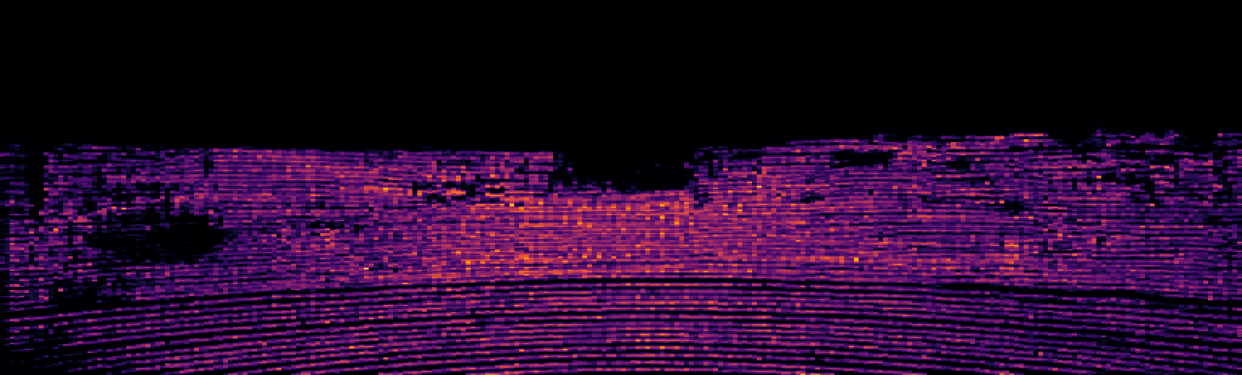}}
		\caption{KITTI Prediction from Combined Image}
			\vspace{1em}
	\end{subfigure}
	%\hfill
	\begin{subfigure}[b]{\linewidth}
		\centering
		{\includegraphics[width=\textwidth]{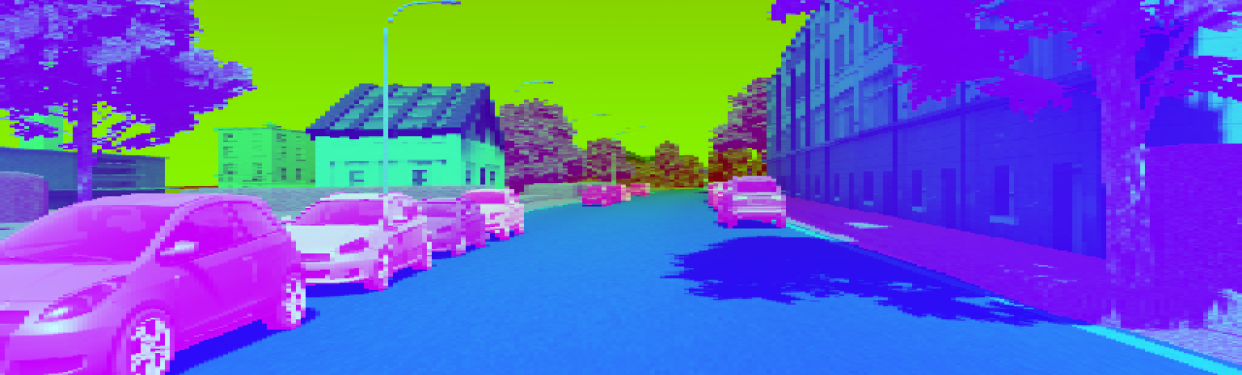}}
	%\caption{Combined \kitti Input Image}	
	\end{subfigure}
	%	\hfill
	\begin{subfigure}[b]{\linewidth}
		\centering
		{\includegraphics[width=\textwidth]{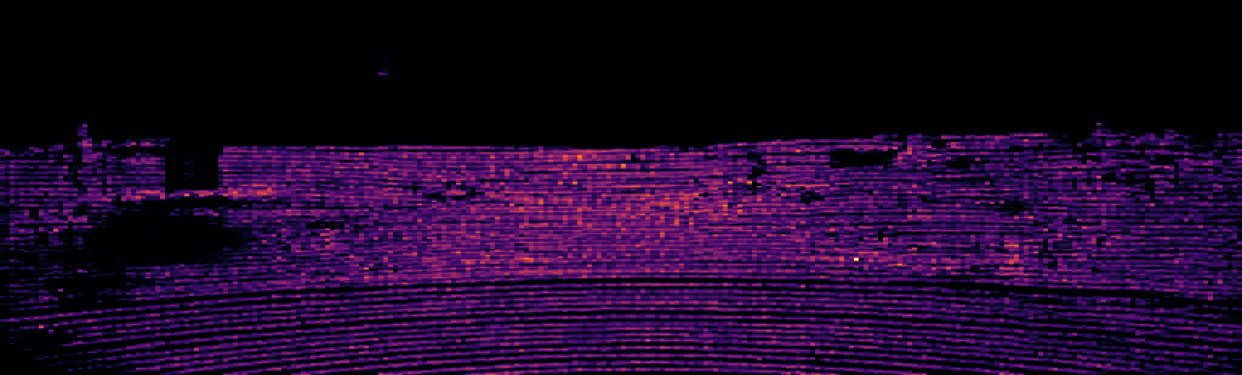}}
		\caption{VKITTI Prediction from Combined Image}
			\vspace{1em}
	\end{subfigure}
	
	\caption{Comparing Results on VKITTI and KITTI. The input image on top of (b) and (c) shows the combined image with RGB (grayscale), depth and segmentation in the color channels.}
	\label{fig:comp_v}
\end{figure}

%\subsubsection{Performance on Synthetic Data}\label{sec:eval_virtual}

For an evaluation of the generalization of our method to synthetic data, we use paired data from \kitti and VKITTI.
We use networks trained on real-world data and apply them to synthetic images from VKITTI.
We predict \lidar images and compare the results with the ground truth \lidar data from \kitti as shown in Fig.~\ref{fig:comp_v}.
Results are shown in Tab.~\ref{tab:example1}.
Results of this comparison strongly depend on the deviation between \kitti and VKITTI data, however, our results still show good correspondence and a behavior comparable to real world data.

\section{\uppercase{Discussion}}
%As the general approach of image-to-image translation is thoroughly evaluated, it may not come as a big surprise that the mapping from camera images or derived representations to 2D \lidar projections is indeed feasible.
%Nonetheless, it is notable that with the presented data and preprocessing, we achieve reasonable performance with a rather short training period.
%We also believe that there is merit to view this simulation problem from a bottom-up as it is complementary step to existing systems that are highly optimized towards a specific configuration.
%After observation of the different \lidar characteristics in our experiments, more complex approaches can focus to improve these individually in the future.
% Considering that the simulation of \lidar sensors, especially with the aid of machine learning techniques, seems to have not have yet received much attention, there are two aspects that should be discussed in further detail. 
% First, the question arises, how the realism of simulated data can be measured.
% Second, we need to think about the potential of learning based algorithms for \lidar simulation in general.

%\subsection{Evaluation and Use Cases of Realistic \lidar Data}

%Yet, it is difficult to compare our results to other works as these often either try to achieve the best possible data simulation instead of a targeting to replicate a certain sensor configuration or train on synthetic data for application on real world data -- exactly the inverse use case of our pipeline.
With our approach of trying to obtain typical and realistic \lidar data, we accept a higher numerical error.
The problem here is that there is no reference for a visual or numerical comparison to existing data based approaches.
The quality of physics based simulation on the other hand depends not only on the implementation of the sensor system but also on the environment.
Even LiDARsim already operates on the aggregated real world data geometry data, which makes a direct result based comparison unfeasible.

The consequence would be to use indirect measures, e.g. follow the sim-to-real approach and select appropriate machine learning algorithms that take point clouds as input and are trained on real data.
This allows to compare the performance between using real and synthetic point cloud sources.
%, in the same manner that a too perfect point cloud could be worse because it creates a bigger domain gap. 
A second option would be to augment training data for such a network with synthetic point clouds to reduce the need for real world data collection.
But
%The issue here is that with synthetic data, it is not clear whether better performance always means more realism and more realism means a smaller domain gap.
%Following the spirit of VKITTI, 
only a high quality virtual twin of an environment would allow a direct comparison. % is also recorded in the form of training data could offer further support.
Especially the geometry ground truth data does not align closely enough in VKITTI (cf.~Fig:~\ref{fig:comp_v}), as the scene is here constructed from a predefined set of objects. 
%This is not to say that further evaluation is not required in this case, 
%Given the complexity of this task, 
We believe that finding ways to better measure realistic simulation will be an important task to achieve virtual testing of \adas.
%Finding effective and practical ways to specifically measure the realism of data based simulations is a very important topic on its own. %\subsection{Image-to-image translation for \lidar Simulation}
%As pix2pix already offers sufficient quality for basic use cases, it seems plausible that more advanced architectures can improve results.

Based on the results of \lidar simulation with pix2pix, experiments with other architectures become very interesting.
However, these often make specific assumptions. % to do so. 
%This does not mean that they would not work for translating between camera images and \lidar projections, but instead,
So there is a need for an architecture that is also specialized for this task and can make better use of combined input data.
%It should be possible to build on existing architectures to create a network that uses segmentation information to account for object type dependent \lidar reconstruction.

On the other hand, the quality of the training data is arguably even more important.
There are developments that can significantly improve this:
Better sensors and thus denser \lidar scans would reduce the resolution discrepancy between camera images and \lidar scans.
%This would enable better training as there is simply more data on the one hand and, specific for our data processing, a better discernibility between gaps in the \lidar resolution and surfaces that are not detectable with \lidar technology.
Similarly, progress in 3d scene reconstruction can improve the data situation as well, but the issues with dynamic objects and movement in general currently can introduce further problems.

%Aside from that, most data sets only store \lidar point clouds or projections from the \lidar perspective into camera perspective.
%Yet, the inverse seems very promising as this avoids the the large number of black pixels in the \lidar projections that arise from the resolution discrepancy.
%The starting point would be an image where each pixel represents a \lidar ray. 

%On the other hand, other types of neural networks like Transformers just to mention one offer further possibilities to process \lidar. \BE{ich wuerde die transformergeschichte entweder begruenden - wieso sollen transformer die speziell geeignet sein (nicht nur evtl besser als convnet) oder den satz loeschen}
Given these possible developments, it is important to have a baseline for future performance comparisons, which our pipeline provides.%and data provides \BE{and data loeschen? wir providen ja keine daten?}.

%The input for the neural network, would then show corresponding color pixels from the camera, as seen from the \lidar perspective. 
%There are also \lidar companies that offer solutions where the camera is directly integrated into the \lidar sensor~\cite{ouster}, which eliminates the need of projecting between different perspectives.
%\BE{this idea is cool, but not sure if it could not raise the question why you didn't do it that way... a critical reviewer could raise it...}

%This format is difficult to obtain as it would require azimuth and polar angle of each point, which is often not supplied. 

%Section, subsection and sub-subsection first paragraph should not
%have the first line indent.

%To remove the paragraph indentation (only necessary for the
%sections), use the command \textit{$\backslash$noindent} before the
%paragraph first word.

%If you use other style files (.sty) you MUST include them in the
%final manuscript zip file.

%\subsubsection{Equations}

%Equations should be placed on a separate line, numbered and
%centered.\\The numbers accorded to equations should appear in
%consecutive order inside each section or within the contribution,
%with the number enclosed in brackets and justified to the right,
%starting with the number 1.

%Example:

%\begin{equation}\label{eq1}
%    a=b+c
%\end{equation}

\section{\uppercase{Conclusion}}
We have presented a pipeline for data based simulation of \lidar sensor behaviour, that generalizes rather well for synthetic data inputs in simulation environments.
While the general quality is not reliable enough for prediction of frame-accurate sensor artefacts, the \lidar simulation is capable of reproducing them in general.
We believe it is sufficient for many use cases, that analyse the performance of \adas functions over many virtual road miles.
%
%Generally, a \lidar simulation like this should be preferable to generic heuristic based \lidar implementations that assume fixed or noise-based distribution of points.
%
%Furthermore, the pipeline serves as an important baseline for \lidar simulation.
%Our initial results can be used for quantitative comparisons of predicting \lidar visibility and the pipeline itself can be enhanced modularly for better performance and be adapted to individual data and simulation needs.
%
%\paragraph{Future Work}
%Going forward, a core task will be to improve the quality of results.
%This requires more elaborate neural network architectures but also ways of enhancing existing data.
With the rapid developments of \lidar technology, better data sets will emerge naturally, while existing data sets can be enhanced with new data fusion techniques to exploit multiple consecutive frames.
%In addition to that, algorithms to better merge available data or possibly combine synthetic with real data could further enhance the upper limit of learning based sensor data generation.
%Additionally, we want to analyse how our pipeline performs in real world applications.
%As our approach can rapidly be applied to other sensor configurations and data sets, we also plan to offer a collection of pretrained, ready-to-use networks and integrations for simulation environments.
%As our approach can rapidly be applied to other sensor configurations and data sets, we also plan to offer a collection of pretrained, ready-to-use networks and integrations for simulation environments.

%The second direction concerns the way we use the sensor data generated by our framework.
%In this paper, we mainly address the  intentional generation of impaired data.
%While there is still further potential, an alternative would be to evaluate our generated point clouds for use in object detection and adapt the approach towards generating more complete data. 

\label{sec:conclusion}

%\vfill
\section*{\uppercase{Acknowledgements}}
Richard Marcus was supported by the Bayerische Forschungsstiftung (Bavarian Research Foundation) AZ-1423-20.
%If any, should be placed before the references section
%without numbering. To do so please use the following command:
%\textit{$\backslash$section*\{ACKNOWLEDGEMENTS\}}
\bibliographystyle{apalike}
{\small
\bibliography{main}}

\begin{thebibliography}{}

\bibitem[Automotive, 2020]{ipg_carmaker_2020}
Automotive, I. (2020).
\newblock {CarMaker}.
\newblock Publisher: IPG Automotive GmbH.

\bibitem[Cabon et~al., 2020]{cabon2020vkitti2}
Cabon, Y., Murray, N., and Humenberger, M. (2020).
\newblock Virtual kitti 2.

\bibitem[Dosovitskiy et~al., 2017]{dosovitskiy_carla_2017}
Dosovitskiy, A., Ros, G., Codevilla, F., Lopez, A., and Koltun, V. (2017).
\newblock {CARLA}: {An} {Open} {Urban} {Driving} {Simulator}.
\newblock {\em arXiv:1711.03938 [cs]}.
\newblock arXiv: 1711.03938.

\bibitem[Gaidon et~al., 2016]{gaidon2016virtual}
Gaidon, A., Wang, Q., Cabon, Y., and Vig, E. (2016).
\newblock Virtual worlds as proxy for multi-object tracking analysis.

\bibitem[Goodfellow et~al., 2014]{goodfellow2014generative}
Goodfellow, I.~J., Pouget-Abadie, J., Mirza, M., Xu, B., Warde-Farley, D.,
  Ozair, S., Courville, A., and Bengio, Y. (2014).
\newblock Generative adversarial networks.

\bibitem[Isola et~al., 2018]{isola_image--image_2018}
Isola, P., Zhu, J.-Y., Zhou, T., and Efros, A.~A. (2018).
\newblock Image-to-{Image} {Translation} with {Conditional} {Adversarial}
  {Networks}.
\newblock {\em arXiv:1611.07004 [cs]}.
\newblock arXiv: 1611.07004.

\bibitem[Johnson et~al., 2016]{johnson_perceptual_2016}
Johnson, J., Alahi, A., and Fei-Fei, L. (2016).
\newblock Perceptual {Losses} for {Real}-{Time} {Style} {Transfer} and
  {Super}-{Resolution}.
\newblock {\em arXiv:1603.08155 [cs]}.
\newblock arXiv: 1603.08155.

\bibitem[Kirillov et~al., 2019]{pointrend}
Kirillov, A., Wu, Y., He, K., and Girshick, R.~B. (2019).
\newblock Pointrend: Image segmentation as rendering.
\newblock {\em CoRR}, abs/1912.08193.

\bibitem[LG, 2021]{simulator_lgsvl_nodate}
LG (2021).
\newblock {LGSVL} {Simulator, https://www.lgsvlsimulator.com/}.

\bibitem[Liao et~al., 2021]{Liao2021ARXIV}
Liao, Y., Xie, J., and Geiger, A. (2021).
\newblock {KITTI}-360: A novel dataset and benchmarks for urban scene
  understanding in 2d and 3d.
\newblock {\em arXiv.org}, 2109.13410.

\bibitem[Manivasagam et~al., 2020]{manivasagam_lidarsim_2020}
Manivasagam, S., Wang, S., Wong, K., Zeng, W., Sazanovich, M., Tan, S., Yang,
  B., Ma, W.-C., and Urtasun, R. (2020).
\newblock {LiDARsim}: {Realistic} {LiDAR} {Simulation} by {Leveraging} the
  {Real} {World}.
\newblock {\em arXiv:2006.09348 [cs]}.
\newblock arXiv: 2006.09348.

\bibitem[NVIDIA, 2021]{nvidia_drivesim}
NVIDIA (2021).
\newblock Nvidia drive sim, https://developer.nvidia.com/drive/drive-sim.

\bibitem[Park et~al., 2019]{park2019SPADE}
Park, T., Liu, M.-Y., Wang, T.-C., and Zhu, J.-Y. (2019).
\newblock Semantic image synthesis with spatially-adaptive normalization.
\newblock In {\em Proceedings of the IEEE Conference on Computer Vision and
  Pattern Recognition}.

\bibitem[Ranftl et~al., 2021]{Ranftl2021}
Ranftl, R., Bochkovskiy, A., and Koltun, V. (2021).
\newblock Vision transformers for dense prediction.
\newblock {\em ArXiv preprint}.

\bibitem[Ronneberger et~al., 2015]{ronneberger2015unet}
Ronneberger, O., Fischer, P., and Brox, T. (2015).
\newblock U-net: Convolutional networks for biomedical image segmentation.

\bibitem[Shah et~al., 2017]{shah_airsim_2017}
Shah, S., Dey, D., Lovett, C., and Kapoor, A. (2017).
\newblock {AirSim}: {High}-{Fidelity} {Visual} and {Physical} {Simulation} for
  {Autonomous} {Vehicles}.
\newblock {\em arXiv:1705.05065 [cs]}.
\newblock arXiv: 1705.05065.

\bibitem[Shorten and Khoshgoftaar, 2019]{shorten_survey_2019}
Shorten, C. and Khoshgoftaar, T.~M. (2019).
\newblock A survey on {Image} {Data} {Augmentation} for {Deep} {Learning}.
\newblock {\em Journal of Big Data}, 6(1):60.

\bibitem[Sun et~al., 2020]{sun_scalability_2020}
Sun, P., Kretzschmar, H., Dotiwalla, X., Chouard, A., Patnaik, V., Tsui, P.,
  Guo, J., Zhou, Y., Chai, Y., Caine, B., Vasudevan, V., Han, W., Ngiam, J.,
  Zhao, H., Timofeev, A., Ettinger, S., Krivokon, M., Gao, A., Joshi, A., Zhao,
  S., Cheng, S., Zhang, Y., Shlens, J., Chen, Z., and Anguelov, D. (2020).
\newblock Scalability in {Perception} for {Autonomous} {Driving}: {Waymo}
  {Open} {Dataset}.
\newblock {\em arXiv:1912.04838 [cs, stat]}.
\newblock arXiv: 1912.04838.

\bibitem[Uhrig et~al., 2017]{Uhrig2017THREEDV}
Uhrig, J., Schneider, N., Schneider, L., Franke, U., Brox, T., and Geiger, A.
  (2017).
\newblock Sparsity invariant cnns.
\newblock In {\em International Conference on 3D Vision (3DV)}.

\bibitem[VTD, 2021]{VTD}
VTD (2021).
\newblock Virtual {Test} {Drive}.

\bibitem[Wang et~al., 2020]{Wang_2020_ACCV}
Wang, L., Goldluecke, B., and Anklam, C. (2020).
\newblock L2r gan: Lidar-to-radar translation.
\newblock In {\em Proceedings of the Asian Conference on Computer Vision
  (ACCV)}.

\bibitem[Wang et~al., 2018a]{wang2018pix2pixHD}
Wang, T.-C., Liu, M.-Y., Zhu, J.-Y., Tao, A., Kautz, J., and Catanzaro, B.
  (2018a).
\newblock High-resolution image synthesis and semantic manipulation with
  conditional gans.
\newblock In {\em Proceedings of the IEEE Conference on Computer Vision and
  Pattern Recognition}.

\bibitem[Wang et~al., 2018b]{wang_pseudo-lidar_2018}
Wang, Y., Chao, W.-L., Garg, D., Hariharan, B., Campbell, M., and Weinberger,
  K.~Q. (2018b).
\newblock Pseudo-{LiDAR} from {Visual} {Depth} {Estimation}: {Bridging} the
  {Gap} in {3D} {Object} {Detection} for {Autonomous} {Driving}.
\newblock {\em arXiv:1812.07179 [cs]}.
\newblock arXiv: 1812.07179.

\bibitem[Weber et~al., 2021]{Weber2021NEURIPSDATA}
Weber, M., Xie, J., Collins, M., Zhu, Y., Voigtlaender, P., Adam, H., Green,
  B., Geiger, A., Leibe, B., Cremers, D., Osep, A., Leal-Taixe, L., and Chen,
  L.-C. (2021).
\newblock Step: Segmenting and tracking every pixel.
\newblock In {\em Neural Information Processing Systems (NeurIPS) Track on
  Datasets and Benchmarks}.

\bibitem[Zhu et~al., 2020]{zhu_unpaired_2020}
Zhu, J.-Y., Park, T., Isola, P., and Efros, A.~A. (2020).
\newblock Unpaired {Image}-to-{Image} {Translation} using {Cycle}-{Consistent}
  {Adversarial} {Networks}.
\newblock {\em arXiv:1703.10593 [cs]}.
\newblock arXiv: 1703.10593.

\end{thebibliography}
\end{document}